\documentclass[10pt,twocolumn,letterpaper]{article}

\usepackage{iccv}
\usepackage{times}
\usepackage{epsfig}
\usepackage{graphicx}
\usepackage{amsmath}
\usepackage{amssymb}
\usepackage{arydshln}
\usepackage[dvipsnames]{xcolor}
\usepackage{soul}
\usepackage{algorithm}
\usepackage{algorithmic}
\usepackage{footnote}
\usepackage{scrextend}
\usepackage{dirtytalk}
\usepackage{amsmath}
\usepackage{multirow}
\usepackage[numbers,sort]{natbib} 
\usepackage[accsupp]{axessibility}  

\makesavenoteenv{tabular}
\makesavenoteenv{table}

\makeatletter
\def\blfootnote{\xdef\@thefnmark{}\@footnotetext}
\makeatother

\usepackage[breaklinks=true,bookmarks=false]{hyperref}

\iccvfinalcopy 

\def\iccvPaperID{7093} 

\ificcvfinal\fi

\newcommand{\methodName}{\textsc{TeachText}}
\newcommand{\methodNameVid}{\textsc{TeachVideo}}

\usepackage[dvipsnames]{xcolor}

\begin{document}

\title{\methodName: CrossModal Generalized Distillation for Text-Video Retrieval}

\author{
Ioana Croitoru \textsuperscript{1,2,*}
\qquad
Simion-Vlad Bogolin\textsuperscript{1,2,*}
\qquad
Marius Leordeanu\textsuperscript{2,3}
\\
Hailin Jin\textsuperscript{4}
\qquad
Andrew Zisserman\textsuperscript{1}
\qquad
Samuel Albanie\textsuperscript{1,5,$\dagger$}
\qquad
Yang Liu\textsuperscript{1,6,$\dagger$} \\
{\small
\textsuperscript{1}Visual Geometry Group, Univ. of Oxford
\qquad
\textsuperscript{2}Inst. of Mathematics of the Romanian Academy
\qquad
\textsuperscript{3}Univ. Politehnica of Bucharest
} \\
{\small
\textsuperscript{4}Adobe Research
\qquad
\textsuperscript{5}Dept. of Engineering, Univ. of Cambridge
\qquad
\textsuperscript{6}Wangxuan Inst. of Computer Technology, Peking Univ.
} \\
}

\maketitle
\ificcvfinal\thispagestyle{empty}\fi

\begin{abstract}
In recent years, considerable progress on the task of text-video retrieval has been achieved by leveraging large-scale pretraining on visual and audio datasets to construct powerful video encoders.
By contrast, despite the natural symmetry, the design of effective algorithms for exploiting large-scale language pretraining remains under-explored.
In this work, we are the first to investigate the design of such algorithms and propose a novel generalized distillation method,~\methodName, which leverages complementary cues from multiple text encoders to provide an enhanced supervisory signal to the retrieval model. Moreover, we extend our method to video side modalities and show that we can effectively reduce the number of used modalities at test time without compromising performance. Our approach advances the state of the art on several video retrieval benchmarks by a significant margin and adds no computational overhead at test time. Last but not least, we show an effective application of our method for eliminating noise from retrieval datasets. Code and data can be found at \url{https://www.robots.ox.ac.uk/~vgg/research/teachtext/}.

\end{abstract}

\section{Introduction}
\label{sec:intro}

\blfootnote{
   \textsuperscript{*}Equal contribution.
   \textsuperscript{$\dagger$}Corresponding authors.
}

The focus of this work is \textit{text-video retrieval}---the task of identifying which video among a pool of candidates best matches a natural language query describing its content. Video search has a broad range of applications across domains such as wildlife monitoring, security, industrial process monitoring and entertainment.  Moreover, as humanity continues to produce video at ever-increasing scale, the ability to perform such searches effectively and efficiently takes on critical commercial significance to video hosting platforms such as YouTube. 

\begin{figure}
    \centering
    \includegraphics[clip,width=0.9\linewidth]{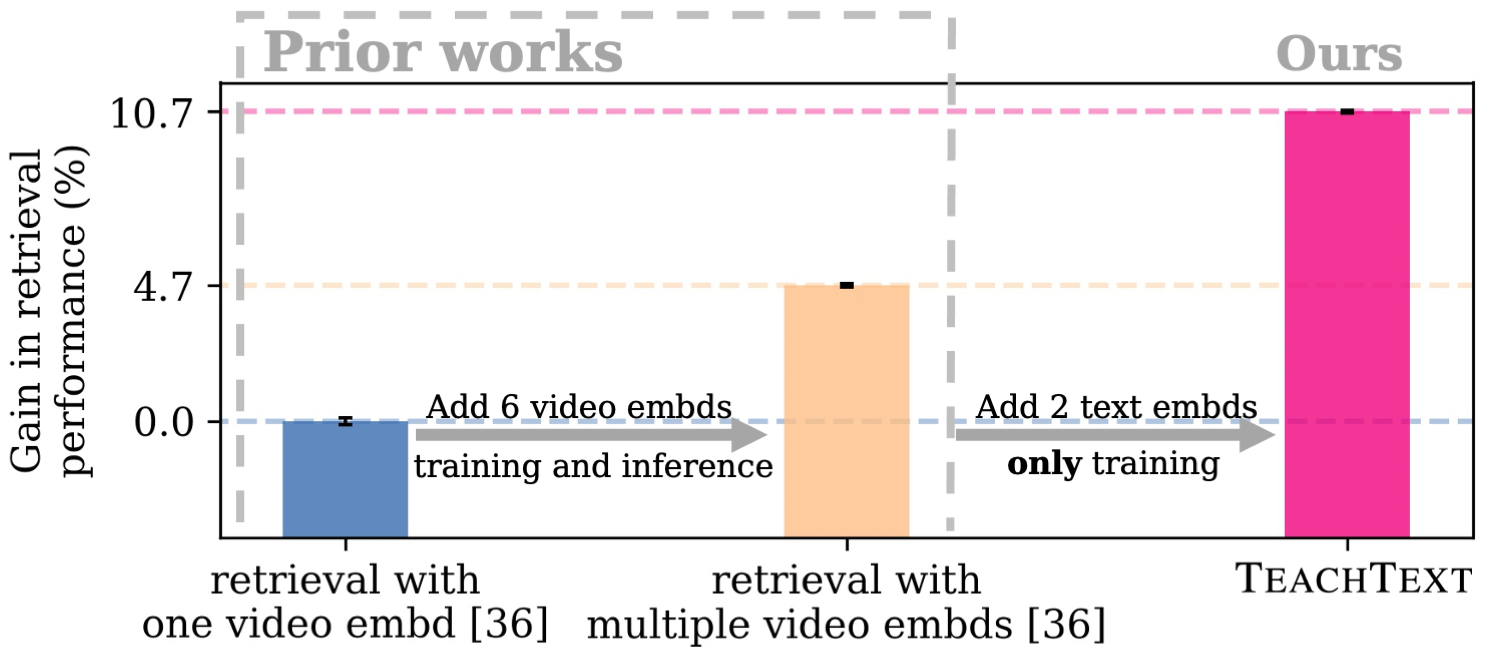}
    \caption{\textbf{Distilling the knowledge from multiple text encoders for stronger text-video retrieval.} Prior works~\cite{miech2018learning,liu2019use,gabeur2020multi} have shown the considerable benefit of transitioning from video encoders that ingest a single modality (\textit{left}) to multi-modal video encoders (\textit{centre}). In this work, we show that retrieval performance can be further significantly enhanced by learning from multiple text encoders through the \methodName{} algorithm which imposes no additional cost during inference. Text-to-video retrieval performance gain (geometric mean of R1-R5-R10) is reported for a~\cite{liu2019use} model as well as for our method on the MSR-VTT \cite{xu2016msr} dataset.
    }
    \mbox{}\vspace{-0.6cm} \\
    \label{fig:teaser}
\end{figure}

A central theme of recently proposed retrieval methods has been the investigation of how to best use multiple video modalities to improve performance. In particular, architectures based on mixtures-of-experts~\cite{miech2018learning,liu2019use} and multi-modal transformers~\cite{gabeur2020multi} have shown the benefit of making use of diverse sets of pre-trained models for related tasks (such as image classification, action recognition and ambient sound classification) as a basis for video encoding during training and testing.

In this work, we explore whether commensurate gains could be achieved by leveraging multiple text embeddings learned on large-scale written corpora. Different from video embeddings using multiple modalities and pretraining tasks, it is less obvious that there is sufficient diversity among collections of text embeddings to achieve a meaningful boost in performance. In fact, 
our inspiration stems from a careful investigation of the performance of different text embeddings across a range of retrieval benchmarks (Fig.~\ref{fig:text-embedding-influence}).  Strikingly, we observe not only that there is considerable variance in performance across text embeddings, but also that \textit{their ranking is not consistent}, strongly supporting the idea of using multiple text embeddings.

Motivated by this finding, we propose a simple algorithm, \methodName{}, to effectively exploit the knowledge captured by collections of text embeddings. 
Our approach requires a \say{student} model to learn from a single or multiple \say{teacher} retrieval models with access to different text embeddings by distilling their text-video similarity matrices into an enhanced supervisory signal. 
As shown in Fig.~\ref{fig:teaser}, \methodName{} is capable of delivering a significant performance gain. Moreover, this gain is complementary to that of adding more video modalities to the video encoder but importantly, unlike the addition of video modalities, does not incur additional computational cost during inference.

Our main contributions can be summarised as follows: 
(1) We propose the \methodName{} algorithm, which leverages the additional information given by the use of multiple text encoders; (2) We show that directly learning the retrieval similarity matrix between the joint query video embeddings, which to the best of our knowledge is novel, is an effective generalized distillation technique for this task (and we compare our approach to alternatives among prior work such as uni-modal relationship distillation~\cite{park2019relational});
(3) We show an application of our approach in eliminating noise from modern training datasets for the text-video retrieval task;
(4) We demonstrate the effectiveness of our approach empirically, achieving state of the art performance on six text-video retrieval benchmarks.

\section{Related Work}
\label{sec:related}

\textbf{Video retrieval methods.} The task of indexing video content to enable retrieval has a rich history in computer vision---sophisticated systems have been developed to find specific objects~\cite{sivic2003video}, actions~\cite{laptev2007retrieving}, predefined semantic categories~\cite{jiang2007towards}, irregularities~\cite{boiman2007detecting} and near-duplicates~\cite{Chum07,shang2010real}. In this work, we focus on the task of retrieving content that matches a given natural language description. For this particular task, there has been considerable interest in developing cross-modal methods that employ a joint-embedding space for text queries and video content~\cite{aytar2008utilizing,xu2015jointly,dong2016word2visualvec,mithun2018learning,yu2018joint,wray2019fine,bain2021frozen}.  These joint video-text embeddings, which aim to map videos and text descriptions into a common space such that matching video and text pairs are close together, form an attractive computational model for tackling this problem, since they allow for efficient indexing (although hierarchical embeddings have also been investigated~\cite{chen2020fine}). Recently, two key themes have emerged towards improving the quality of these embeddings. First, large-scale weakly supervised pretraining methods~\cite{miech2019howto100m,miech2019end,korbar2020video} have sought to expand their training data by exploiting the speech contained in the videos themselves as a supervisory signal. Second, the integration of multiple modalities (which has long been considered important for semantic indexing~\cite{snoek2005multimodal}) has been shown to yield significant gains in performance~\cite{miech2018learning,liu2019use,gabeur2020multi,patrick2020support}.  We focus on candidates from this latter theme as a basis for investigating our approach. 

\textbf{Text embeddings.}
The representation of language through learned embeddings has been widely studied~\cite{mikolov2013efficient, radford2018improving, radford2019language} and applied in a variety of natural language processing applications. Several works have demonstrated that even with large-scale pretraining, there still are benefits to finetuning the models on the target task~\cite{radford2018improving,devlin2019bert} and that larger models (often employing multiple attention heads) yield higher performance~\cite{devlin2019bert}.
Recently, \cite{burns2019language} provided a detailed comparisons on the importance of language features for vision applications and proposes a word embedding that is specifically designed for vision tasks. In this work, we first study how various pretrained language embeddings affect the performance for text-video retrieval and then propose a method to take advantage of the benefits of combining multiple text embeddings.

\textbf{Knowledge Distillation/Privileged Information.} The purpose of knowledge distillation is to transfer knowledge from one model (teacher) to another model (student). This idea was originally introduced in the context of decision tree simplification~\cite{breiman1996born} and model compression~\cite{bucilua2006model}, and later extended by~\cite{hinton2015distilling} who formalised this knowledge transfer as the temperature-parameterised process of \textit{knowledge distillation}.  The concept was further generalised in the unifying framework of \textit{generalized distillation}~\cite{lopez2016unifying} for learning with privileged information~\cite{vapnik2009new} (via \textit{similarity control} and \textit{knowledge transfer}~\cite{vapnik2015learning}), together with knowledge distillation~\cite{hinton2015distilling}. Our approach distills knowledge of the similarities between video and text samples into the student and therefore represents a form of generalized distillation. While most knowledge distillation methods train the student with the teacher's outputs as targets, more recent methods propose different approaches~\cite{romero2014fitnets, zagoruyko2016paying, huang2017like}. Of most relevance to our approach, ~\cite{park2019relational} transfer mutual relations of data examples and propose distance-wise and angle-wise distillation losses that penalize structural differences in relations instead of training the student to mimic the output of the teacher---we compare to their approach in Sec.~\ref{sec:experiments}.

\section{Motivation and intuition}
\label{sec:motivation}

\begin{figure*}
    \centering
    \includegraphics[width=\linewidth]{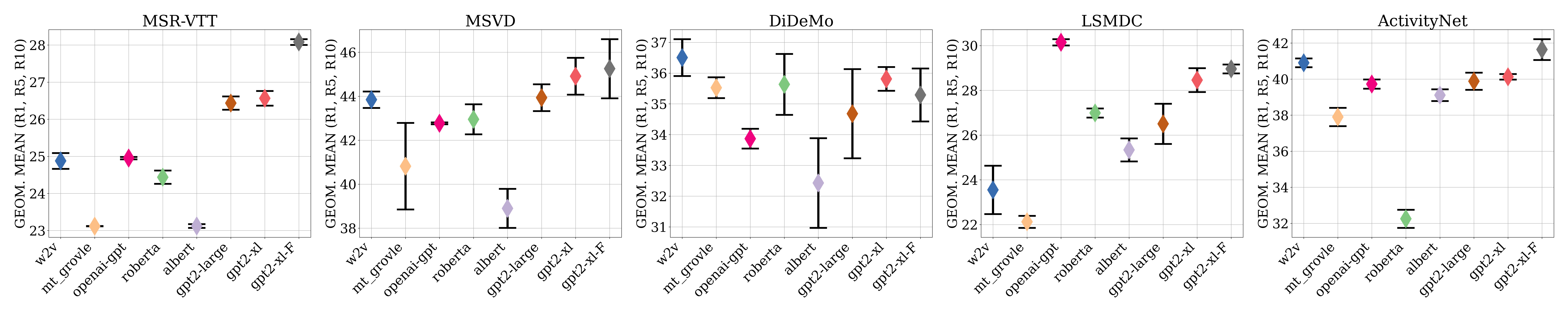}
    \caption{\textbf{Influence of varying the text embedding.} Different text embeddings are presented on the x axis: w2v~\cite{mikolov2013efficient}, mt\_grovle~\cite{burns2019language}, openai-gpt~\cite{radford2018improving}, roberta~\cite{liu2019roberta}, albert~\cite{lan2019albert}, gpt2-large~\cite{radford2019language}, gpt2-xl~\cite{radford2019language}, gpt2-xl-F along with their performance in geometric mean of $R1$-$R5$-$R10$ on five datasets. For each
experiment, we report the mean (diamond) and standard deviation (error bar) of three randomly seeded runs.This study is performed using the CE retrieval architecture~\cite{liu2019use}: each model differs only in its use of pre-trained text embedding at input. We observe a significant variance in performance when changing the text embedding, both across and within datasets. The difference in rankings across datasets suggests the presence of additional information among different text embeddings.
    }
    \mbox{}\vspace{-0.7cm} \\
    \label{fig:text-embedding-influence}
\end{figure*}

\begin{figure}
    \centering
    \includegraphics[width=0.9\linewidth]{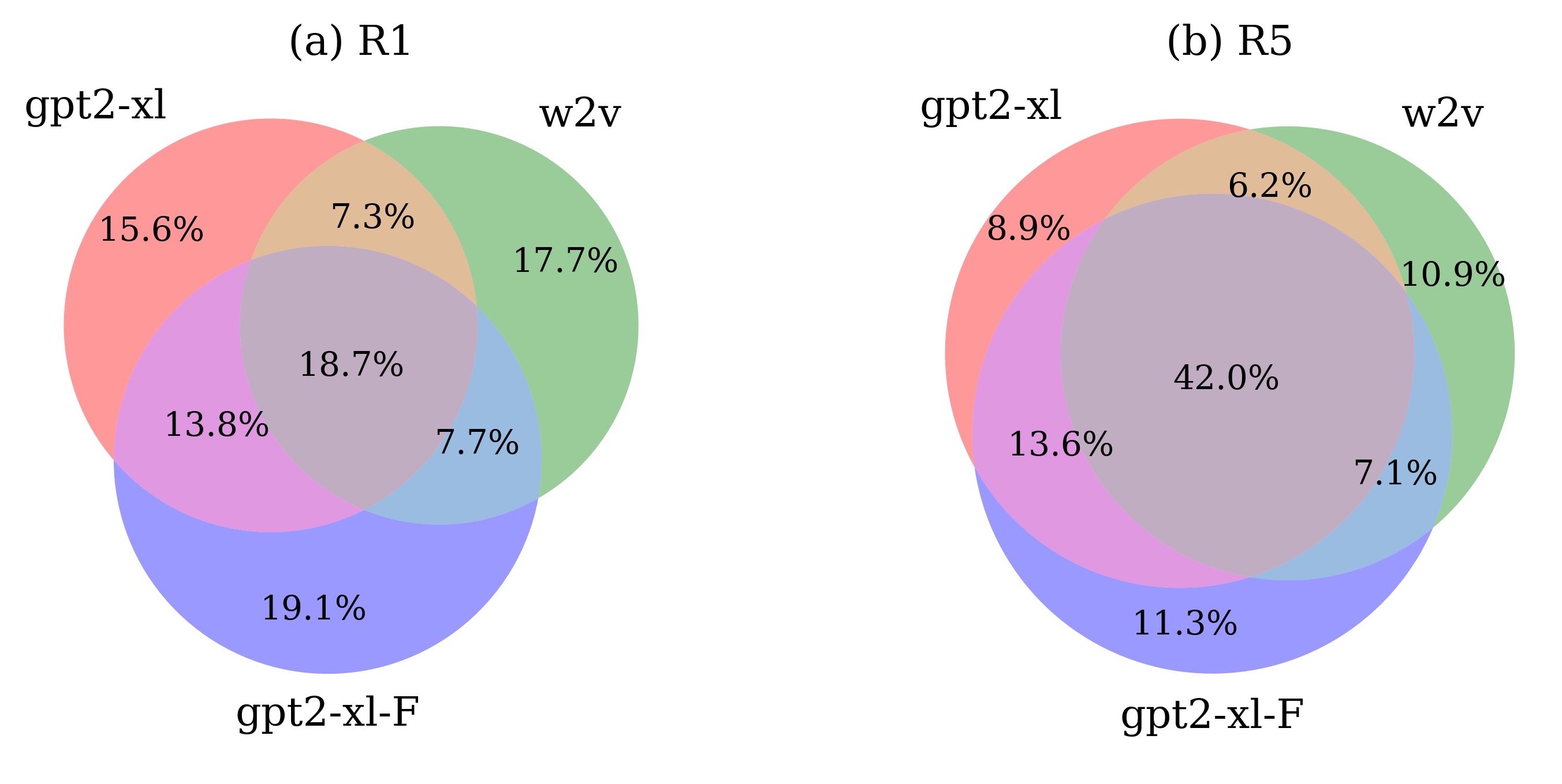}
    \caption{\textbf{Share of correctly retrieved samples based on the used pre-trained text embedding on MSR-VTT.} We observe that each embedding has a considerable share of sample retrieved correctly only by itself (in terms of R1 left and R5 right), further justifying our approach. Best viewed in color.
    }
    \mbox{}\vspace{-0.7cm} \\
    \label{fig:venn-text-embds}
\end{figure}

Recently, \cite{radford2019language} points out that even though language representation learning systems (such as~\cite{liu2019roberta,lan2019albert,radford2018improving}) are pre-trained on vast amounts of data, they are still sensitive to slight changes in the data distribution and task specification. In this way, most systems can be viewed as \textit{narrow experts rather than competent generalists}.

Consequently, in Fig.~\ref{fig:text-embedding-influence} we investigate how the usage of different off-the-shelf pre-trained text embeddings affects the retrieval performance. We observe that there is significant variance both within and across datasets, suggesting that each embedding captures different types of information. Our intuition is that this information comes from the diversity of architectures, pretraining datasets and pretraining objectives, which differs across the text embeddings. 

Next, we give details about the used text embeddings and summarise the key differences between them in relationship with our findings. Word2vec (w2v)~\cite{mikolov2013efficient} is a lightweight text embedding that is widely used for vision tasks~\cite{NEURIPS2020_3493894f,cao2017deep,wang2019describing}. Multi-task GrOVLE (mt\_grovle)~\cite{burns2019language}, is an extension of w2v that is specially designed for vision-language tasks (in our experiments, however, we find that it slightly under-performs w2v).
The finetuned transformer language model (openai-gpt)~\cite{radford2018improving} embedding is trained on a book corpus containing long stretches of contiguous text. We observe that it performs well on datasets that have longer text queries such as ActivityNet. RoBERTa and ALBERT~\cite{liu2019roberta,lan2019albert} are based on the BERT architecture~\cite{devlin2019bert} and are trained on the same data which consists of unpublished books and Wikipedia articles. RoBERTa~\cite{liu2019roberta} focuses on hyperparameter optimization and shows that greater model capacity 
leads to better performance while  ALBERT\cite{lan2019albert} proposes some parameter-reduction techniques to reduce memory consumption and increase training speed. In our experiments, we observe a high variation in performance when comparing the two. In contrast to the other embeddings, gpt2\cite{radford2019language} is trained on a crawled dataset that was designed to be as diverse as possible. We observe that gpt2 performs most robustly in our experiments, especially on smaller datasets such as MSR-VTT and MSVD. However, it nevertheless exhibits a domain gap to each corpus (highlighted by the fact that performance increases when fine-tuning gpt2-xl, termed gpt2-xl-F throughout the paper, on queries from the text-video retrieval datasets). 

Additionally, in Fig.~\ref{fig:venn-text-embds} we show how many correctly retrieved queries are shared between three text embeddings on MSR-VTT: gpt2-xl, gpt2-xl-F and w2v. Only around 19\% (R1), respectively 42\% (R5) queries are correctly retrieved by all the three considered text embeddings. This means that a significantly number of queries are sensitive to the used text embedding, consolidating our intuition.

\section{Method}
\label{sec:method}

\setlength{\textfloatsep}{0.5cm}
\setlength{\dbltextfloatsep}{0.7cm}

\begin{figure*}
    \centering
    \includegraphics[width=\linewidth]{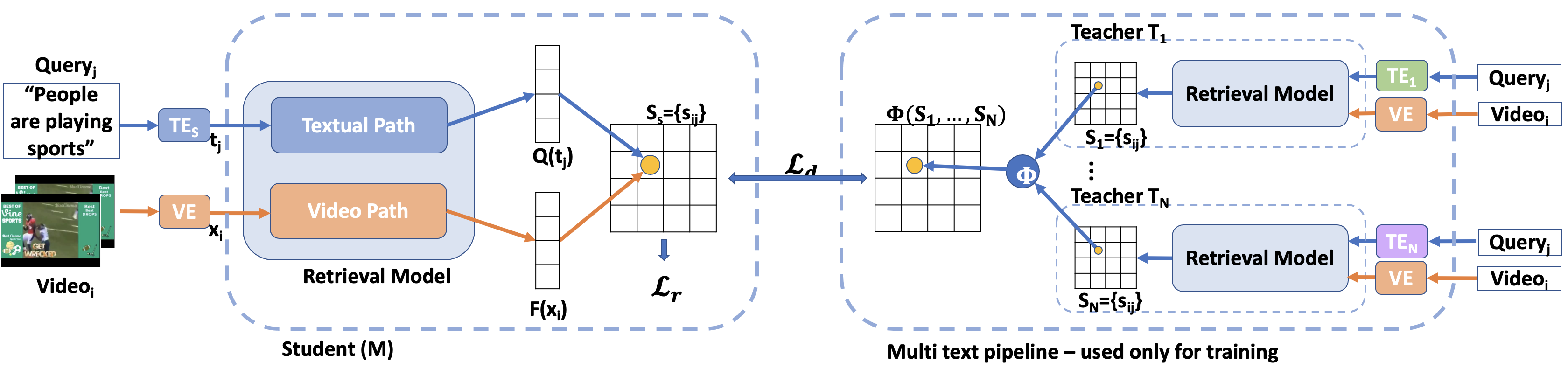}
    \caption{\textbf{\methodName{} teacher-student framework overview.} Given a batch of input videos and queries in natural language during training, the student model, $M$ (left) and teacher models $T_1, \dots, T_N$ (right) each produce similarity matrices (visualised as square grids). The similarity matrix produced by $M$ is encouraged to match the aggregated matrices of the teachers through the distillation loss $\mathcal{L}_d$ in addition to the retrieval loss $\mathcal{L}_r$. Note that both the student and teachers ingest the same video embeddings (VE), but employ different text embeddings (TE$_S$ for the student, TE$_1,\dots,$TE$_N$ for the teachers). At test time, the teacher models are discarded.
    }
    \mbox{}\vspace{-0.7cm} \\
    \label{fig:method}
\end{figure*}

Motivated by the findings from Sec.~\ref{sec:motivation}, our work aims to study the influence of using multiple text embeddings for text-video retrieval. 

\subsection{Problem description and learning setup}
\label{sec:pb-desc}

Let $D=\{(v_i, c_i)\}_{i=1}^n$ be a dataset of paired videos and captions. Following the multi-modal experts approach of~\cite{miech2018learning,liu2019use,gabeur2020multi}, for each video we have access to a collection of video embeddings (sometimes referred to as \say{experts}) $x_i$ extracted from the various modalities of video $v_i$ using a pretrained video encoder (\textit{VE}) in addition to a text embedding $t_i$ (extracted using a text encoder, \textit{TE}) for each caption/query $c_i$\footnote{These embeddings are produced by models that have been trained on relevant tasks (such as action recognition for the video encoder and language modelling for the text encoder)}.  The objective of the text-video retrieval task is to learn a model $M(x_i, t_j)$ which assigns a high similarity value to pairings $(x_i, t_j)$ of video and text embeddings that are in correspondence (i.e. $i = j$) and a low similarity otherwise. As is common in the literature~\cite{bojanowski2015weakly,miech2018learning}, we parameterise the model as a dual-encoder that produces joint-embeddings in a shared space such that they can be compared directly $M(x_i, t_j) = F(x_i)^T Q(t_j) \in \mathbb{R}$ where $F$ and $Q$ represent the learnt video and text encoder respectively.
To train the video and text encoder for the task of retrieval, we adopt a contrastive ranking loss~\cite{socher2014grounded}:
\vspace{-0.25cm}
\begin{equation}
\begin{aligned}
    \mathcal{L}_r = \frac{1}{B}\sum_{i=1}^{B}\sum_{i \neq j} \left[ max(0, s_{ij} - s_{ii} +m) + \right. \\  \left. max(0, s_{ji} - s_{ii} + m)  \right]
    \label{eqn:ranking}
\end{aligned}
\mbox{}\vspace{-0.15cm} \\
\end{equation}

where $B$ represents the batch size used during training, $s_{ij} = F(x_i)^TQ(t_j)$ is the similarity score between the encoded video $F(x_i)$ and query $Q(t_j)$ while $m$ is the margin.

The key idea behind our approach is to learn a retrieval model, $M$, that, in addition to the loss described above, also has access to information provided by a collection of pre-trained \say{teacher} retrieval models which are trained on the same task but ingest different text embeddings.

\subsection{\methodName{} algorithm}
\label{sec:teachtext}

To enhance the retrieval performance of model $M$, we propose the \methodName{} algorithm which aims to exploit cues from multiple text embeddings.
An overview of our approach is provided in Fig.~\ref{fig:method}. Initially, we train a collection of teacher models $\{T_k: k \in \{1, \dots, N\}\}$ for the text-video retrieval task using the approach described in Sec.~\ref{sec:pb-desc}. The teachers share the same architecture but each model $T_k$ uses a different text embedding as input (extracted using a pre-trained text encoder \textit{TE}$_k$). In the second phase the parameters of the teachers are frozen.  We then proceed by sampling a batch of $B$ pairs of videos and captions and computing a corresponding similarity matrix $S_k \in \mathbb{R}^{B \times B}$ for each teacher $T_k$ (Fig.~\ref{fig:method} right).  These $N$ similarity matrices are then combined with an aggregation function, $\Phi: \mathbb{R}^{N \times B\times B} \rightarrow \mathbb{R}^{B\times B}$, to form a single supervisory similarity matrix (Fig.~\ref{fig:method}, centre-right). Concurrently, the batch of videos and captions are likewise processed by the student model, $M$, which produces another similarity matrix, $S_s \in \mathbb{R}^{B \times B}$. Finally, in addition to the standard retrieval loss (Eq.~\ref{eqn:ranking}), a distillation loss, $\mathcal{L}_d$, encourages the $S_s$ to lie close to the aggregate $\Phi(S_1, \dots, S_N)$.
The algorithm is summarized in Alg.~\ref{alg:teach-text}.
During inference, the teacher models are discarded and the student model $M$ requires only a single text embedding.
Next, we give details of the distillation loss used for the similarity matrix learning.

\begin{algorithm}
\caption{\methodName{} algorithm}
\label{alg:teach-text}
\begin{algorithmic}[1]
\STATE \textbf{Phase 1:} \textit{Learn teacher models}
\begin{ALC@g}
\STATE Train $N$ teacher models $T_k=(F_k, Q_k)$, $k \in \{1, \dots, N\}$ using the training pairs $(x_i, t_j^k)$ where $t_j^k$ represents the text modality used by teacher $T_k$ in a standard retrieval training setup (Sec.~\ref{sec:pb-desc}).
\end{ALC@g}
\STATE \textbf{Phase 2:} \textit{Learn the student model}, $M=(F, Q)$
\begin{ALC@g}
\FOR  {minibatch of $B$ paired samples $\{(v_i, c_i)\}$}

\STATE For each pair $(v_i, c_i)$ extract video experts and text embedding pairs $(x_i, t_i)$ using \textit{VE} and $TE_S$.
\STATE Compute student similarity matrix $S_s$ where $S_s(i,j) = F(x_i)^TQ(t_j)$ for $i,j \in \{1,\dots,B\}$
\STATE Compute the loss $\mathcal{L}_r$ via Eqn.~\ref{eqn:ranking} using $S_s$.

\FOR{teacher $T_k$, $k=1,\dots,N$} 

\STATE For each pair $(v_i, c_i)$ extract the video experts and text embedding pairs $(x_i, t_i^k)$ using $VE$ and $TE_k$.

\STATE Compute the similarity matrix $S_k$ where $S_k(i,j) = F_k(x_i)^TQ_k(t_i^k)$ for $i,j\in \{1,\dots,B\}$.
\ENDFOR

\STATE Compute aggregate teacher matrix $\Phi(S_1, \dots, S_N)$. 

\STATE Compute the loss $\mathcal{L}_d$ between $S_s$ and $\Phi(S_1, \dots, S_N)$ via Eqn.~\ref{eqn:distil}.
\STATE Update $M$ with gradients computed from the composite loss $\mathcal{L} =\mathcal{L}_r + \mathcal{L}_d$.
\ENDFOR
\end{ALC@g}
\end{algorithmic}
\end{algorithm}

\subsection{Learning the similarity matrix}
\label{sec:sim-mat}

As noted in Sec.~\ref{sec:pb-desc}, the essence of the retrieval task is to create a model that is able to establish cross-modal correspondences between videos and texts/queries, assigning a high similarity value to a pairing in which a query accurately describes a video, and a low similarity otherwise. This renders the similarity matrix a rich source of information about the knowledge held by the model. In order to be able to transfer knowledge from the teachers to the student, we encourage the student to produce a similarity matrix that matches an aggregate of those produced by the teachers. In this way, we convey information about texts and video correspondences without strictly forcing the student to produce exactly the same embeddings as the teachers.
To this end, we define the similarity matrix distillation loss as:
\vspace{-0.2cm}
\begin{equation}
    \mathcal{L}_d = \frac{1}{B} \sum_{i=1}^{B}\sum_{j=1}^{B} \left[ \textit{l}(\Phi(i, j),  S_s(i, j)) \right]
    \label{eqn:distil}
\mbox{}\vspace{-0.2cm} \\
\end{equation}

where $B$ represents the batch size,  $\Phi = \Phi(S_1, \dots, S_N)$ represents the aggregate of the teacher similarity matrices and $S_s$ represents the similarity matrix of the student.
Finally, inspired from other distillation works such as~\cite{park2019relational}, $\textit{l}$ represents the Huber loss and is defined as
\vspace{-0.2cm}
\begin{equation}
    \textit{l}(x,y) = 
    \begin{cases}
    \frac{1}{2} (x-y)^2 \quad \quad \text{if } |x-y| \leq 1, \\
    |x-y| - \frac{1}{2} \quad \quad otherwise
    \end{cases}
\mbox{}\vspace{-0.2cm} \\
\end{equation}

We explored several forms of aggregation function and found that a simple element-wise mean, $\Phi(S_1, \dots, S_N) = \frac{1}{N}\sum_{k=1}^{N}S_k$, worked well in practice.

The idea of learning directly the cross-modal similarity matrix is, to the best of our knowledge novel. It draws inspiration from the work of relational knowledge distillation~\cite{park2019relational} which considered the idea of learning from relationships and introduced two algorithms to implement this concept in a uni-modal setting through pairwise and triplet distance sampling. We compare our matrix learning approach with theirs in Sec.~\ref{sec:experiments}.

\subsection{Student model}
\label{sec:student}

A key advantage of our approach is that it is agnostic to the architectural form of the student and teachers, and thus the student (and teachers) can employ any method 
from the current literature. We test our \methodName{} algorithm using three different recent works MoEE~\cite{miech2018learning}, CE~\cite{liu2019use}, MMT~\cite{gabeur2020multi} as the student and teacher base architectures. All these works employ multi-modal video encoders for the text-video retrieval task. 
For more details, please consult the original paper of each method.

\noindent \textit{Establishing a stronger baseline.} In addition to these models, we also investigate our approach on a model which shares the CE architecture of~\cite{liu2019use} but includes a series of small technical improvements to provide a stronger baseline against which we also test the \methodName{} algorithm.
Starting from this base architecture, we refine the input embedding selection, finding that the face and OCR video modalities employed by~\cite{liu2019use} do not consistently produce improvement so we remove them as inputs to the video encoder. We update the model to use the more powerful gpt2-xl text embedding of~\cite{radford2019language} and following~\cite{gabeur2020multi}, we finetune this text embedding on captions from the target dataset to bring additional improvement. Combining all of these changes (ablations provided in Sec.~\ref{sec:ablations} and Fig.~\ref{fig:technical-improvements}a) results in the CE+ model which we include as an additional baseline. 
Thus, in summary we use four (\cite{miech2018learning,liu2019use,gabeur2020multi} and CE+) different base architectures for the student model.

\subsection{Teacher models}
\label{sec:teacher}

The teacher models use the same architecture as the student model. Concretely, for each of the four base architectures described in Sec.~\ref{sec:student}, we create a pool of multiple teachers, each using a different pre-trained text embedding as input.
The candidate text embeddings we consider are: mt\_grovle~\cite{burns2019language}, openai-gpt~\cite{radford2018improving}, gpt2-large~\cite{radford2019language}, gpt2-xl~\cite{radford2019language}, w2v~\cite{mikolov2013efficient}. So, we obtain a set of up to five models that form the teachers $T_k$, $k=1..5$ used by \methodName{}.

\subsection{Training and implementation details}
\label{sec:impl}

In order to train our final student, we combine the retrieval loss and the proposed distillation loss $\mathcal{L} = \mathcal{L}_r + \mathcal{L}_d$. Our model is trained in Pytorch~\cite{paszke2019pytorch} using the Adam~\cite{kingma2014adam} optimizer. \methodName{} does not add any additional trainable parameters or modalities to the final model. Moreover, when
training the student using \methodName{}, only the additional loss term $\mathcal{L}_d$ is added, all other hyper-parameters remaining the
same.
\section{Experimental setup}
\label{sec:experiments}

\begin{figure}[b]
    \centering
    \includegraphics[width=\linewidth]{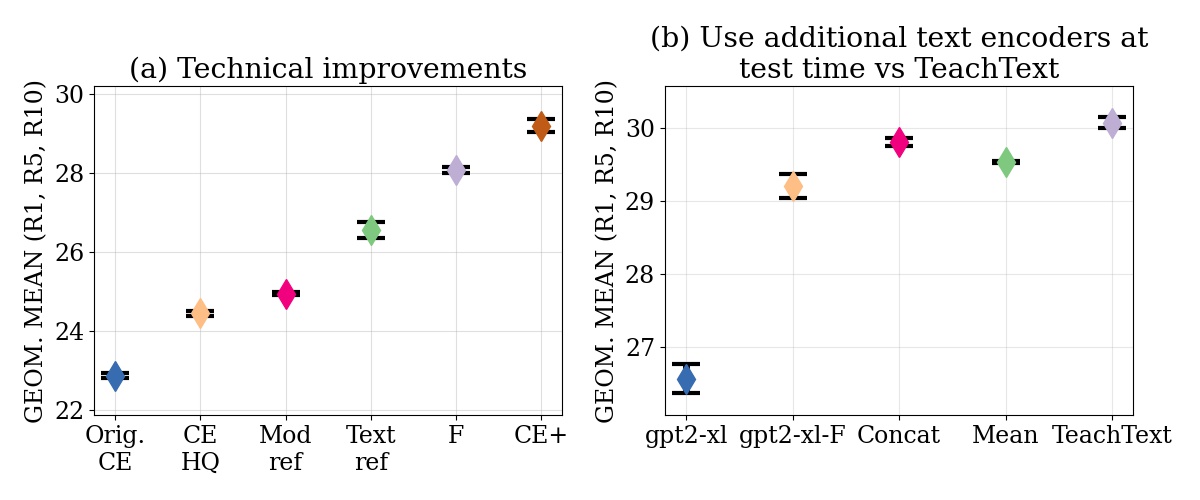}
    \caption{\textbf{(a) Baseline improvements}. The y-axes (scaled for clarity) denotes retrieval performance on MSR-VTT. We begin by presenting the performance of the original CE~\cite{liu2019use}. Firstly, we correct compression artefacts in the pre-processing used for embedding extraction (\textit{CE HQ}, more details in Suppl. Mat.). Secondly, we refine the used video modalities and text modalities (\textit{Mod ref} and \textit{Text ref}). Finally, we finetune the text embedding (\textit{F}) and change the optimizer to Adam~\cite{kingma2014adam}, thus obtaining the \textit{CE+} baseline. 
    \textbf{(b) Use additional text embeddings at inference time}.
    All experiments were performed with the same architecture~\cite{liu2019use}, but with different text embeddings: gpt2-xl (first bullet), gpt2-xl-F (second bullet), the concatenation of gpt2-xl and gpt2-xl-F (third bullet), the mean of gpt2-xl and gpt2-xl-F (fourth bullet) and using \methodName{} (last bullet). By using multiple text embeddings at test time, which introduces an overhead, a boost in performance is obtained. However, by using \methodName{} there is no additional overhead at test time and the performance is superior.}
    \mbox{}\vspace{-0.7cm} \\
    \label{fig:technical-improvements}
\end{figure}

\subsection{Datasets description}
To provide an extensive comparison we test our approach on seven video datasets that have been explored in recent works as benchmarks for the task of text-video retrieval: \textbf{LSMDC}~\cite{rohrbach2017movie}, \textbf{DiDeMo}~\cite{anne2017localizing},
\textbf{MSVD}~\cite{chen2011collecting}, \textbf{MSRVTT}~\cite{xu2016msr}, \textbf{ActivityNet}~\cite{caba2015activitynet}, \textbf{VaTeX}~\cite{wang2019vatex} and \textbf{QuerYD}~\cite{oncescu20queryd}. We follow the same experimental setup as prior works~\cite{liu2019use,gabeur2020multi,chen2020fine,patrick2020support}.

\subsection{Metrics}
To assess performance, we follow prior work  (e.g ~\cite{dong2016word2visualvec,miech2018learning,mithun2018learning,yu2018joint,miech2019howto100m, liu2019use, gabeur2020multi}) and report standard retrieval metrics, including R@K (recall at rank K, where higher is better) and MdR (median rank where lower is better).  For certain analyses, to maintain conciseness we report the geometric mean of R@1, R@5 and R@10 rather than individual metrics (this statistic aims to be representative of overall retrieval performance). The numbers are reported for the task of retrieving a video given text queries \texttt{t2v} which is more common in real world applications. The numbers for the reverse task \texttt{v2t} and the number of parameters for each model are reported in the Suppl. Mat. For each
experiment, we report the mean and standard deviation of three randomly seeded runs.

\subsection{Ablations}
\label{sec:ablations}
In this section we present an extensive study of our proposed approach. Following the setup used in prior works~\cite{liu2019use, gabeur2020multi} we conduct ablations on the MSR-VTT dataset~\cite{xu2016msr}, except where otherwise stated.

\begin{table*}
\begin{center}
\resizebox{\linewidth}{!}{
\begin{tabular}{c|cc|cc|cc|cc|cc|cc}%
\hline%
\hline%
\multirow{2}{*}{Model}&\multicolumn{2}{c|}{MSRVTT}&\multicolumn{2}{c|}{MSRVTT 1k-A}&\multicolumn{2}{c|}{MSVD}&\multicolumn{2}{c|}{DiDeMo}&\multicolumn{2}{c|}{LSMDC}&\multicolumn{2}{c}{ActivityNet}\\%
&Base&\methodName{}&Base&\methodName{}&Base&\methodName{}&Base&\methodName{}&Base&\methodName{}&Base&\methodName{}\\%
\hline%
MoEE&$24.4_{\pm0.1}$&$\mathbf{25.8}_{\pm0.1}$&$41.6_{\pm0.4}$&$\mathbf{43.4}_{\pm0.6}$&$41.8_{\pm0.3}$&$\mathbf{43.2}_{\pm0.5}$&$33.2_{\pm1.4}$&$\mathbf{40.2}_{\pm0.7}$&$23.8_{\pm0.4}$&$\mathbf{26.0}_{\pm0.5}$&$40.1_{\pm0.3}$&$\mathbf{45.2}_{\pm0.1}$\\%
CE&$24.4_{\pm0.1}$&$\mathbf{25.9}_{\pm0.1}$&$42.0_{\pm0.8}$&$\mathbf{43.8}_{\pm0.3}$&$42.3_{\pm0.6}$&$\mathbf{42.6}_{\pm0.4}$&$34.2_{\pm0.4}$&$\mathbf{39.5}_{\pm0.5}$&$23.7_{\pm0.3}$&$\mathbf{25.5}_{\pm0.5}$&$40.4_{\pm0.3}$&$\mathbf{45.0}_{\pm0.6}$\\
MMT&-&-&$44.7_{\pm0.4}$&$\mathbf{45.6}_{\pm0.7}$&-&-&-&-&$24.6_{\pm0.7}$&$\mathbf{25.9}_{\pm0.6}$&$44.0_{\pm0.4}$&$\mathbf{47.9}_{\pm0.4}$\\
CE+&$29.2_{\pm0.2}$&$\mathbf{30.4}_{\pm0.0}$&$50.3_{\pm0.2}$&$\mathbf{50.9}_{\pm0.4}$&$46.5_{\pm1.0}$&$\mathbf{46.6}_{\pm0.5}$&$35.8_{\pm0.4}$&$\mathbf{40.4}_{\pm0.4}$&$28.1_{\pm0.3}$&$\mathbf{30.7}_{\pm0.3}$&$39.7_{\pm0.0}$&$\mathbf{46.3}_{\pm0.2}$\\

\hline%
\end{tabular}}
\end{center}
\vspace{-0.2cm}
\caption{\textbf{Method generality}. Retrieval performance (geometric mean of R1-R5-R10) on various datasets when applying \methodName{} on top of different base models: MoEE\cite{miech2018learning}, CE\cite{liu2019use}, MMT\cite{gabeur2020multi} (on available datasets) and CE+. We present in bold cases where \methodName{} brings an improvement over the base architecture. We observe that our method improves the performance for all underlying base models and on all datasets.
\label{tab:generality}}
\vspace{-0.5cm}
\end{table*}

\indent \textbf{Baseline improvements.}
We propose CE+ as an additional baseline which consists of a series of technical improvements to the model of ~\cite{liu2019use}. As seen in Fig.~\ref{fig:technical-improvements}a each modification described in Sec.~\ref{sec:student} brings additional gain over the base architecture. We observe in particular that finetuning the text embedding on the target dataset has a high influence, further highlighting the critical role played by text embeddings and justifying their study.
In addition to other changes we found that certain video embedding expert features were highly sensitive to compression choices used in video pre-processing, which we correct accordingly (more details in Suppl. Mat.). Please note that for a fair comparison, in Sec.~\ref{sec:sota-comparison} we report the numbers of re-training the methods~\cite{miech2018learning,liu2019use} using these embeddings extracted with the updated pre-processing which yields a higher performance than the ones reported in the original papers.\\
\indent \textbf{Using multiple text embeddings during inference.}
\methodName{} makes no use of additional information at test time. However, it is natural to ask whether the additional text embeddings can be trivially included as part of the model architecture. In Fig.~\ref{fig:technical-improvements}(b) we compare our approach with some relatively simple text embedding aggregation techniques, which require access to multiple text embeddings during both training and inference. We observe that \methodName{} outperforms these aggregation techniques such as direct concatenation or mean of the text embeddings,
suggesting that the proposed method is effective in capturing the additional information given by multiple text embeddings. Moreover, the text encoder of existing systems~\cite{miech2018learning,liu2019use,gabeur2020multi} typically employs many parameters, so adding multiple text embeddings to the architecture adds a significant number of parameters (100M+). For example, the concatenation of two text embeddings (provided that they have the same size) almost doubles the total number of parameters for CE+. In contrast, when employing \methodName{}, no parameters are added.\\ 
\indent \textbf{Teacher variation.}
\begin{figure}
    \centering
    \includegraphics[width=\linewidth]{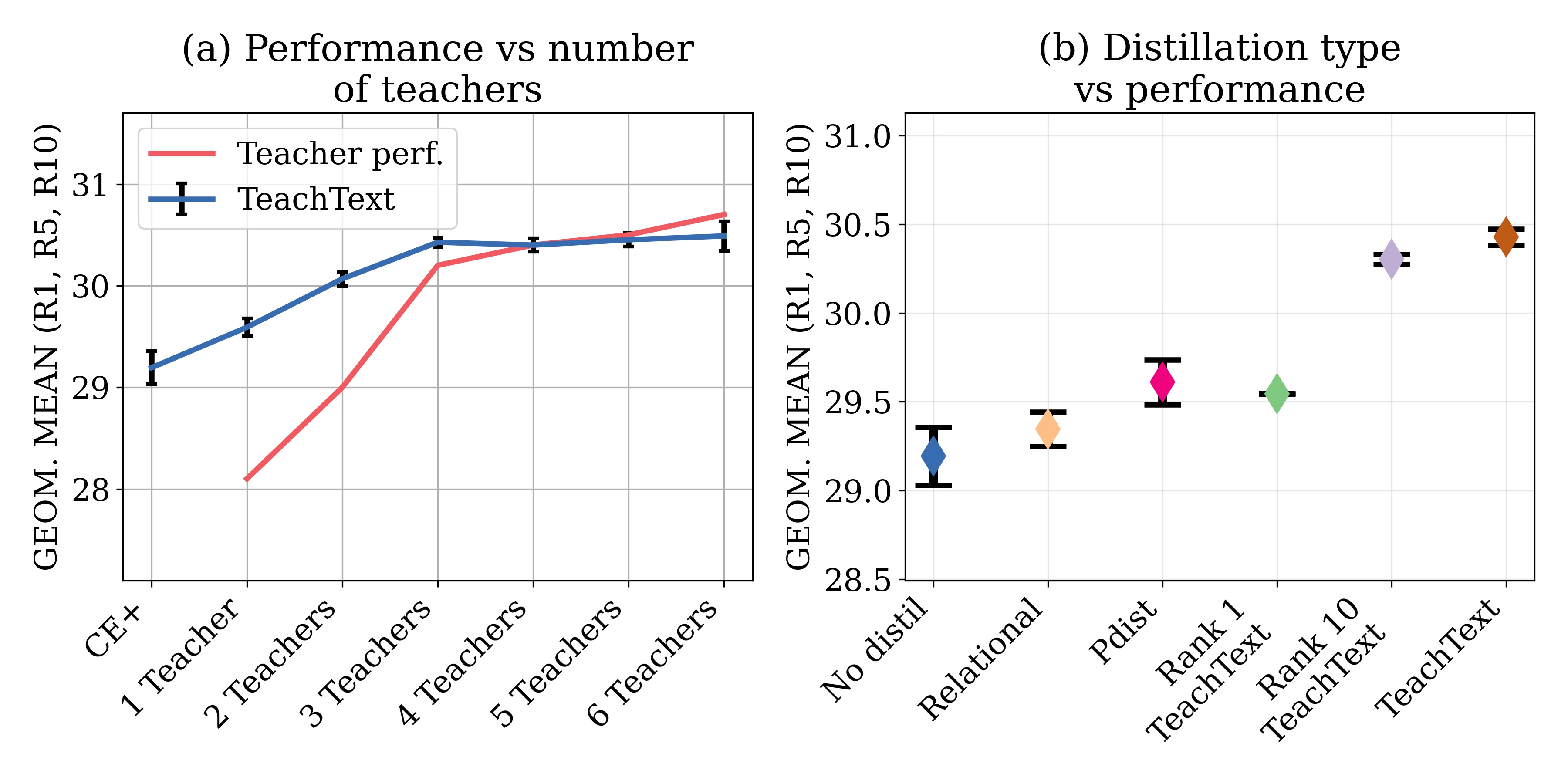}
    \caption{\textbf{(a) Teacher study.} We show the influence of learning from different number of teachers on the MSR-VTT dataset (all students share the same CE+ model, y-axes scaled for clarity). The teachers were added in the following order: gpt2-xl, w2v, gpt2-xl-F, mt\_grovle, openai-gpt, gpt2-large. The performance of the combined teachers grows as more teachers are added, however it reaches a plateau after the first 3 teachers. The trend is similar for student performance. \textbf{(b) Distillation type.} Presenting various alternatives for distilling the information from the teacher: relational distillation~\cite{park2019relational} which preserves intra-text and intra-video relationships, pairwise distance distillation (\textit{Pdist} - adapting~\cite{park2019relational} for cross modal relationships), ranking distillation inspired by \cite{tang2018ranking} at Rank 1 and Rank 10 and \methodName{}. The first bullet represents the student without distillation.
    }
    \mbox{}\vspace{-0.3cm} \\
    \label{fig:teacher-study+distil-type}
\end{figure}
The teacher models share the same architecture with the student, but use a different text embedding. We next conduct an ablation on the influence of the number of used teachers. We observe in Fig.~\ref{fig:teacher-study+distil-type}a that performance increases with the addition of more teachers. Since the combined performance of the teachers after adding more than 3 remains about the same, we do not obtain a further improvement. Thus, for our final experiments presented in Sec.~\ref{sec:sota-comparison} we use a combination of three teachers, having the following text embeddings: w2v~\cite{mikolov2013efficient}, gpt2-xl~\cite{radford2019language} and gpt2-xl-F (gpt2-xl finetuned on the captions from the target dataset). A study of how each individual text embedding affects the final performance can be found in the Suppl. Mat. section \textit{Teacher study}, where we observe that even when using a teacher with lower performance (w2v), the student has a significant boost in performance. \\
\indent\textbf{Distillation ablation.}
\begin{figure}
    \centering
    \includegraphics[width=\linewidth]{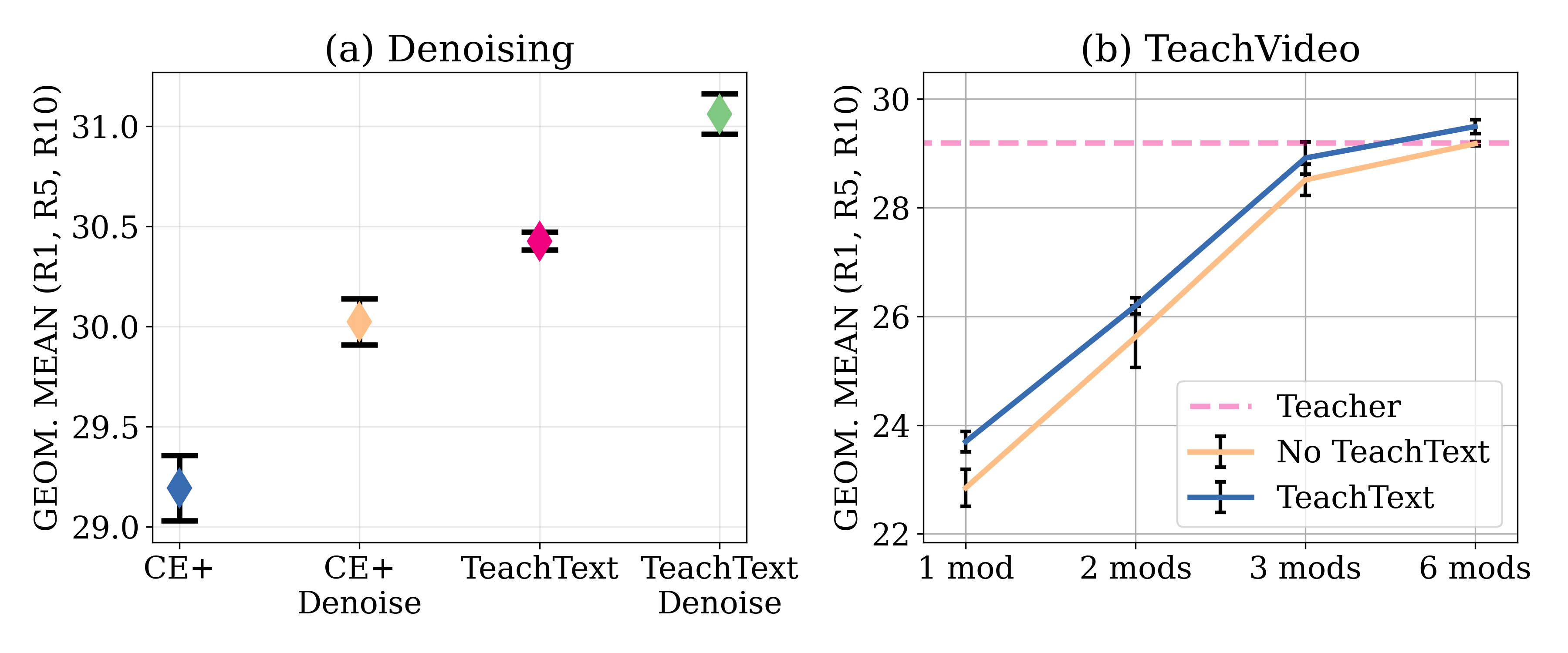}
    \caption{\textbf{(a) Denoising.} We present the effect of denoising on retrieval performance on MSR-VTT (y-axes scaled for clarity). Some of the captions available in datasets with multiple captions per video may be noisy and actively harm the training process. We estimate the degree of noise present in a caption by looking at the teacher rank and drop the caption if necessary. We observe the effectiveness of denoising when applied in isolation (\textit{CE+} vs \textit{CE+ Denoise}) and in conjunction with the full \methodName{} method. \textbf{(b) \methodNameVid{} - Extension to video side modalities.} We observe that our method can be effective in taking advantage of the additional information brought by using multiple video side modalities, without incurring computational overhead at test time.}
    \label{fig:denoising+teachvideo}
    \mbox{}\vspace{-0.6cm} \\
\end{figure}
We compare the proposed learning of the similarity matrix with other distillation alternatives. As seen in Fig.\ref{fig:teacher-study+distil-type}b, our proposed approach is effective in capturing the relationships between video and text.
We first provide comparisons between \methodName{} and several possible instantiations of relational distillation~\cite{park2019relational}. Indeed, given the highly general nature of~\cite{park2019relational}, \methodName{} can be interpreted within this framework as a particular relational configuration that employs cross-modal distillation through batches of similarity matrices.
Since the original work of~\cite{park2019relational} considered single-modality applications, we explore two variations of~\cite{park2019relational} as baselines for the text-video retrieval task. The first one (Relational), preserves the same intra-text and intra-video relationships independently. We use the same cost function as in~\cite{park2019relational} and enforce it on both video and text embeddings. The second approach (Pdist), uses the cross modal pairwise distances as a relation measure between text and video as opposed to the similarity matrix. While these methods indeed bring a gain, we observe that \methodName{} is more effective. \\
\indent We also provide a baseline inspired by the work of~\cite{tang2018ranking} which highlights the importance of looking only at the top K predictions given by the teacher. To do so, we enforce the same similarities using \methodName{} only for the top K ranks given by the teacher rather than for the whole mini-batch. We show the performance for K=1 and K=10 (\textit{Rank 1} and \textit{Rank 10} presented in Fig.\ref{fig:teacher-study+distil-type}b). Restricting to only top K predictions when distilling the similarity matrix results in a slight drop in performance.\\
\indent \textbf{Method generality.}
To demonstrate the generality of \methodName, we test it against three state of the art methods~\cite{miech2018learning,liu2019use,gabeur2020multi} in addition to the proposed CE+ baseline. In Tab.~\ref{tab:generality} we observe a consistent gain in performance, independent of the base architecture. Moreover, a gain is achieved across all the datasets that we tested, having over 5\% absolute gain on DiDeMo and ActivityNet datasets for MoEE, CE and CE+ models. Note that for MMT~\cite{gabeur2020multi} we report results on the datasets included in the public implementation provided by the authors\footnote{\url{https://github.com/gabeur/mmt}}.\\
\indent \textbf{Method application -- Denoising.}
One immediate application of our method is data denoising. Existing real-world text-video datasets for the retrieval task suffer from label noise which can harm training. More concretely, in crowd-sourced datasets such as MSR-VTT there are some captions that are highly ambiguous/generic (e.g "A tutorial is presented", "Clip showing different colours", "A man is writing") and can describe multiple videos from the dataset.
We therefore propose to use \methodName{} teachers to filter out such cases. For this scenario, we simply remove low-ranked predictions given by teachers and re-train the student using only the new samples. Specifically, we remove all sentences for which the correct video is not ranked in top 40 from the training set.
This method is best-suited for datasets where multiple captions per video are available, ensuring that we can remove noisy captions without removing the video itself from training. Following this, we apply the denoising on MSR-VTT and MSVD datasets with the CE+ model. As seen in Fig.~\ref{fig:denoising+teachvideo}a, this can be an effective way of further improving the results. Please note, denoising is not used in any other ablations. \\
\indent \textbf{\methodNameVid{} -- Extension to video modalities.}
While the focus of this work is the use of multiple text embeddings, it is natural to consider whether this approach can be extended to the video encoder modalities. Thus, we introduce the \methodNameVid{} algorithm which follows the same setup as the original \methodName{}, but now the teacher has access to multiple video modalities instead of multiple text modalities. In this study, all students and all teachers use the same text embedding, so we can assess the gains due to \methodNameVid{}. By employing \methodNameVid{} we retain the computational advantage of requiring fewer video modalities during inference. As it can be seen from our experiments presented in Fig.~\ref{fig:denoising+teachvideo}b, the method is effective and brings a boost over the original student.
We believe this extension may be useful in scenarios in which limited computational resources are available during inference. 

Qualitative examples and other ablation studies are presented in Suppl. Mat.

\begin{table}
\begin{center}
\resizebox{0.9\linewidth}{!}{
\begin{tabular}{c|c|c|c|c}%
\hline%
\hline%
Model&$R@1\uparrow$&$R@5\uparrow$&$R@10\uparrow$&$MdR\downarrow$\\%
\hline%
Dual\cite{dong2019dual}&$7.7$&$22.0$&$31.8$&$32.0$\\%
HGR\cite{chen2020fine}&$9.2$&$26.2$&$36.5$&$24.0$\\%
MoEE\cite{miech2018learning}\footnote{Please note that the numbers reported are higher than in the original paper due to compression artefacts correction.\label{hq-note}}&$11.1_{\pm0.1}$&$30.7_{\pm0.1}$&$42.9_{\pm0.1}$&$15.0_{\pm0.0}$\\%
CE\cite{liu2019use}\footref{hq-note} &$11.0_{\pm0.0}$&$30.8_{\pm0.1}$&$43.3_{\pm0.3}$&$15.0_{\pm0.0}$\\%
\hline%
TT-CE&$11.8_{\pm0.1}$&$32.7_{\pm0.1}$&$45.3_{\pm0.1}$&$13.0_{\pm0.0}$\\%
TT-CE+&$\textbf{15.0}_{\pm0.1}$&$\textbf{38.5}_{\pm0.1}$&$\textbf{51.7}_{\pm0.1}$&$\textbf{10.0}_{\pm0.0}$\\%

\hline%
\end{tabular}}
\end{center}
\vspace{-0.2cm}
\caption{\textbf{MSR-VTT full split: Comparison to state of the art.} \label{tab:msrvtt-final-sota}}
\vspace{-0.3cm}
\end{table}

\begin{table}
\begin{center}
\resizebox{0.9\linewidth}{!}{
\begin{tabular}{c|c|c|c|c}%
\hline%
\hline%
Model&$R@1\uparrow$&$R@5\uparrow$&$R@10\uparrow$&$MdR\downarrow$\\%
\hline%
MoEE\cite{miech2018learning}\footref{hq-note}&$21.6_{\pm1.0}$&$50.8_{\pm1.1}$&$65.6_{\pm0.7}$&$5.3_{\pm0.6}$\\
CE\cite{liu2019use}\footref{hq-note}&$21.7_{\pm1.3}$&$51.8_{\pm0.5}$&$65.7_{\pm0.6}$&$5.0_{\pm0.0}$\\
MMT\cite{gabeur2020multi}&$24.6_{\pm0.4}$&$54.0_{\pm0.2}$&$67.1_{\pm0.5}$&$4.0_{\pm0.0}$\\%
SSB\cite{patrick2020support}&$27.4$&$56.3$&$67.7$&$\mathbf{3.0}$\\%
\hline%
TT-MMT&$24.8_{\pm0.2}$&$55.9_{\pm0.7}$&$68.5_{\pm1.0}$&$4.3_{\pm0.5}$\\%
TT-CE+&$\mathbf{29.6}_{\pm0.3}$&$\mathbf{61.6}_{\pm0.5}$&$\mathbf{74.2}_{\pm0.3}$&$\mathbf{3.0}_{\pm0.0}$\\%
\hline%
\end{tabular}}
\end{center}
\vspace{-0.2cm}
\caption{\textbf{MSR-VTT 1k-A split\cite{yu2018joint}: Comparison with others.} \label{tab:msrvtt-jsfusion-final-sota}}
\vspace{-0.3cm}
\end{table}

\begin{table}
\begin{center}
\resizebox{0.9\linewidth}{!}{
\begin{tabular}{c|c|c|c|c}%
\hline%
\hline%
Model&$R@1\uparrow$&$R@5\uparrow$&$R@10\uparrow$&$MdR\downarrow$\\%
\hline%
VSE++\cite{faghri2017vse++}&$15.4$&$39.6$&$53.0$&$9.0$\\%
M-Cues\cite{mithun2018learning}&$20.3$&$47.8$&$61.1$&$6.0$\\%
MoEE\cite{miech2018learning}\footref{hq-note}&$21.1_{\pm0.2}$&$52.0_{\pm0.7}$&$66.7_{\pm0.2}$&$5.0_{\pm0.0}$\\%
CE\cite{liu2019use}\footref{hq-note}&$21.5_{\pm0.5}$&$52.3_{\pm0.8}$&$67.5_{\pm0.7}$&$5.0_{\pm0.0}$\\%
\hline%
TT-CE&$22.1_{\pm0.4}$&$52.2_{\pm0.5}$&$67.2_{\pm0.6}$&$5.0_{\pm0.0}$\\%
TT-CE+&$\mathbf{25.4}_{\pm0.3}$&$\mathbf{56.9}_{\pm0.4}$&$\mathbf{71.3}_{\pm0.2}$&$\mathbf{4.0}_{\pm0.0}$\\%
\hline%
\end{tabular}}
\end{center}
\vspace{-0.2cm}
\caption{\textbf{MSVD: Comparison to state of the art methods}. \label{tab:msvd-final-sota}}
\vspace{-0.3cm}
\end{table}

\begin{table}
\begin{center}
\resizebox{0.9\linewidth}{!}{
\begin{tabular}{c|c|c|c|c}%
\hline%
\hline%
Model&$R@1\uparrow$&$R@5\uparrow$&$R@10\uparrow$&$MdR\downarrow$\\%
\hline%
S2VT\cite{venugopalan2014translating}&$11.9$&$33.6$&$-$&$13.0$\\%
FSE\cite{zhang2019lookahead}&$13.9_{\pm0.7}$&$36.0_{\pm0.8}$&$-$&$11.0_{\pm0.0}$\\%
MoEE\cite{miech2018learning}\footref{hq-note}&$16.1_{\pm1.0}$&$41.2_{\pm1.6}$&$55.2_{\pm1.6}$&$8.3_{\pm0.5}$\\%
CE\cite{liu2019use}\footref{hq-note}&$17.1_{\pm0.9}$&$41.9_{\pm0.2}$&$56.0_{\pm0.5}$&$8.0_{\pm0.0}$\\%
\hline%
TT-CE&$21.0_{\pm0.6}$&$47.5_{\pm0.9}$&$61.9_{\pm0.5}$&$\mathbf{6.0}_{\pm0.0}$\\%
TT-CE+&$\mathbf{21.6}_{\pm0.7}$&$\mathbf{48.6}_{\pm0.4}$&$\mathbf{62.9}_{\pm0.6}$&$\mathbf{6.0}_{\pm0.0}$\\%
\hline%
\end{tabular}}
\end{center}
\vspace{-0.2cm}
\caption{\textbf{DiDeMo: Comparison to state of the art methods}. \label{tab:didemo-final-sota}}
\vspace{-0.3cm}
\end{table}

\begin{table}
\begin{center}
\resizebox{0.9\linewidth}{!}{
\begin{tabular}{c|c|c|c|c}%
\hline%
\hline%
Model&$R@1\uparrow$&$R@5\uparrow$&$R@10\uparrow$&$MdR\downarrow$\\%
\hline%
JSFus\cite{yu2018joint}&$9.1$&$21.2$&$34.1$&$36.0$\\%
MoEE\cite{miech2018learning}\footref{hq-note}&$12.1_{\pm0.7}$&$29.4_{\pm0.8}$&$37.7_{\pm0.2}$&$23.2_{\pm0.8}$\\%
CE\cite{liu2019use}\footref{hq-note}&$12.4_{\pm0.7}$&$28.5_{\pm0.8}$&$37.9_{\pm0.6}$&$21.7_{\pm0.6}$\\%
MMT\cite{gabeur2020multi}&$13.2_{\pm0.4}$&$29.2_{\pm0.8}$&$38.8_{\pm0.9}$&$21.0_{\pm1.4}$\\%
\hline%
TT-MMT&$13.6_{\pm0.5}$&$31.2_{\pm0.4}$&$40.8_{\pm0.5}$&$17.7_{\pm0.5}$\\%
TT-CE+&$\mathbf{17.2}_{\pm0.4}$&$\mathbf{36.5}_{\pm0.6}$&$\mathbf{46.3}_{\pm0.3}$&$\mathbf{13.7}_{\pm0.5}$\\%
\hline%
\end{tabular}}
\end{center}
\vspace{-0.2cm}
\caption{\textbf{LSMDC: Comparison to state of the art methods}. \label{tab:lsmdc-final-sota}}
\vspace{-0.3cm}
\end{table}

\begin{table}
\begin{center}
\resizebox{0.9\linewidth}{!}{
\begin{tabular}{c|c|c|c|c}%
\hline%
\hline%
Model&$R@1\uparrow$&$R@5\uparrow$&$R@50\uparrow$&$MdR\downarrow$\\%
\hline%
MoEE\cite{miech2018learning}\footref{hq-note}&$19.7_{\pm0.3}$&$50.0_{\pm0.5}$&$92.0_{\pm0.2}$&$5.3_{\pm0.5}$\\%
CE\cite{liu2019use}\footref{hq-note}&$19.9_{\pm0.3}$&$50.1_{\pm0.7}$&$92.2_{\pm0.6}$&$5.3_{\pm0.5}$\\%
HSE\cite{zhang2018cross}&$20.5$&$49.3$&$-$&$-$\\%
MMT\cite{gabeur2020multi}&$22.7_{\pm0.2}$&$54.2_{\pm1.0}$&$93.2_{\pm0.4}$&$5.0_{\pm0.0}$\\%
SSB\cite{patrick2020support}&$\mathbf{26.8}$&$58.1$&$93.5$&$\mathbf{3.0}$\\%
\hline%
TT-MMT&$25.0_{\pm0.3}$&$\mathbf{58.7}_{\pm0.4}$&$95.6_{\pm0.2}$&$4.0_{\pm0.0}$\\%
TT-CE+&$23.5_{\pm0.2}$&$57.2_{\pm0.5}$&$\mathbf{96.1}_{\pm0.1}$&$4.0_{\pm0.0}$\\%
\hline%
\end{tabular}}
\end{center}
\vspace{-0.2cm}
\caption{\textbf{ActivityNet: Comparison to state of the art methods}. \label{tab:activity-net-final-sota}}
\vspace{-0.3cm}
\end{table}

\begin{table}
\begin{center}
\resizebox{0.9\linewidth}{!}{
\begin{tabular}{c|c|c|c|c}%
\hline%
\hline%
Model&$R@1\uparrow$&$R@5\uparrow$&$R@10\uparrow$&$MdR\downarrow$\\%
\hline%
VSE\cite{kiros2014unifying}&$28.0$&$64.3$&$76.9$&$3.0$\\%
Dual\cite{dong2019dual}&$31.1$&$67.4$&$78.9$&$3.0$\\%
VSE++\cite{faghri2017vse++}&$33.7$&$70.1$&$81.0$&$2.0$\\%
HGR\cite{chen2020fine}&$35.1$&$73.5$&$83.5$&$2.0$\\%
SSB\cite{patrick2020support}&$44.6$&$81.8$&$89.5$&$\mathbf{1.0}$\\%
CE\cite{liu2019use}&$47.9_{\pm0.1}$&$84.2_{\pm0.1}$&$91.3_{\pm0.1}$&$2.0_{\pm0.0}$\\%
\hline%
TT-CE&$49.7_{\pm0.1}$&$85.6_{\pm0.1}$&$92.4_{\pm0.1}$&$2.0_{\pm0.0}$\\%
TT-CE+&$\mathbf{53.2}_{\pm0.2}$&$\mathbf{87.4}_{\pm0.1}$&$\mathbf{93.3}_{\pm0.0}$&$\mathbf{1.0}_{\pm0.0}$\\%
\hline%
\end{tabular}}
\end{center}
\vspace{-0.2cm}
\caption{\textbf{VaTeX: Comparison to state of the art methods}. \label{tab:vatex-final-sota}}
\vspace{-0.3cm}
\end{table}

\begin{table}
\begin{center}
\resizebox{0.9\linewidth}{!}{
\begin{tabular}{c|c|c|c|c}%
\hline%
\hline%
Model&$R@1\uparrow$&$R@5\uparrow$&$R@10\uparrow$&$MdR\downarrow$\\%
\hline%
MoEE\cite{miech2018learning}&$11.6_{\pm1.3}$&$30.2_{\pm3.0}$&$43.2_{\pm3.1}$&$14.2_{\pm1.6}$\\%
CE\cite{liu2019use}&$13.9_{\pm0.8}$&$37.6_{\pm1.2}$&$48.3_{\pm1.4}$&$11.3_{\pm0.6}$\\%
\hline%
TT-CE&$14.2_{\pm1.4}$&$36.6_{\pm2.0}$&$\mathbf{51.1}_{\pm2.1}$&$\mathbf{9.7}_{\pm1.2}$\\%
TT-CE+&$\mathbf{14.4}_{\pm0.5}$&$\mathbf{37.7}_{\pm1.7}$&$50.9_{\pm1.6}$&$9.8_{\pm1.0}$\\%
\hline%
\end{tabular}}
\end{center}
\vspace{-0.2cm}
\caption{\textbf{QuerYD: Comparison to state of the art methods}. \label{tab:queryd-final-sota}}
\vspace{-0.3cm}
\end{table}

\subsection{Comparison to prior work}
\label{sec:sota-comparison}

As it can be seen in Tab.\ref{tab:msrvtt-final-sota},\ref{tab:msrvtt-jsfusion-final-sota},\ref{tab:msvd-final-sota},\ref{tab:didemo-final-sota},\ref{tab:lsmdc-final-sota},\ref{tab:activity-net-final-sota},\ref{tab:vatex-final-sota},\ref{tab:queryd-final-sota} our approach is effective and achieves state of the art results on six datasets. All methods are trained for the retrieval task using only the samples from the target datasets. In order to be as fair as possible, we included the results of our \methodName{} (abbreviated TT) applied also to the best existing method for each dataset. So, the architecture and the used features are identical during inference (e.g. TT-CE has the same architecture and uses the same video and text embeddings as CE). We highlight in bold the best performing method.

\section{Conclusion}
\label{sec:conclusion}

In this paper, we present a novel algorithm \methodName \ for the text-video retrieval task. We use a teacher-student paradigm where a student learns to leverage the additional information given by one or multiple teachers, sharing the architecture, but each using a different pre-trained text embedding at input. In this way, we achieve state of the art results on six benchmarks. Finally, we present an application of our approach for denoising video retrieval datasets.
\vspace{-0.2cm}
\\
\noindent\textbf{Acknowledgements.} This work was supported by EPSRC Programme Grants Seebibyte
EP/M013774/1 and VisualAI EP/T028572/1, and a gift from Adobe. M.L.\ was supported by UEFISCDI, under project EEA-RO-2018-0496. The authors would like to thank Gyungin Shin and Iulia Duta for assistance. S.A.\ would like to acknowledge the support of Z.\ Novak and S.\ Carlson in enabling his contribution.

{\small
\bibliographystyle{ieee_fullname}
\bibliography{egbib}
}
\clearpage
\appendix
\begin{center}
    \textbf{\Large Appendix}
\end{center}
\vspace{0.5cm}

In this supplementary material, we provide additional details on the video embeddings (Sec.~\ref{supp:experts}) and text embeddings (Sec.~\ref{supp:text-embed}) used in the main submission. We provide further details of Fig. 1 from the main paper (Sec.~\ref{supp:teaser-details}) as well as details on optimization (Sec.~\ref{supp:opt-details}), modifications to the embedding pre-processing pipeline used in prior work (Sec.~\ref{supp:correction}) and summaries of the datasets used (Sec.~\ref{supp:dataset}). Finally, we include additional ablations (Sec.~\ref{supp:ablations}) and a more comprehensive set of metrics for comparison with previous work, along with qualitative results (Sec.~\ref{supp:sota}).

\section{Video embeddings (experts) description}
\label{supp:experts}
In this work, we used the set of pretrained experts considered by the authors of~\cite{liu2019use}.  For completeness, we summarise here the manner in which these experts were extracted.

\begin{itemize}
    \item Two form of action experts are used: \textit{Action(KN)} and \textit{Action(IG)}. The former is an I3D architecture trained on Kinetics~\cite{carreira2017quo}, which produces 1024-dimensional embeddings from frame clips extracted at 25fps and center cropped to 224 pixels.  The \textit{Action(IG)} model is a 34-layer R(2+1)D model~\cite{tran2018closer} that has ben trained on IG-65m~\cite{ghadiyaram2019large}: it operates on frames extracted at 30 fps in clips of 8 at 112 $\times$ 112 pixel resolution.
    \item Two forms of object experts are used, named \textit{Obj(IN)} and \textit{Obj(IG)}.  They are produced from frame-level embeddings extracted at 25fps.  The \textit{Obj(IN)} model consists of an SENet-154 backbone~\cite{hu2019squeeze} which has been trained on ImageNet for image classification.   \textit{Obj(IG)} is formed from a ResNext-101~\cite{xie2017aggregated} extractor which was trained on Instagram data that was weakly labelled with hashtags~\cite{mahajan2018exploring}.  For both models, frames are resized to 224$\times$224 pixels.
    \item The face expert uses a ResNet50~\cite{he2016identity} that has been trained for task of face classification on the VGGFace2 dataset~\cite{Cao18}, producing a 512-dimensional embedding for each detected face following detection. 
    \item The audio expert is produced using the VGGish model, trained for audio classification on the YouTube-8m dataset and described by \cite{hershey2017}. 
    \item The scene expert is a 2208-dim embedding that is extracted frames (at 25 fps) for a center crop of 224$\times$224 pixels.  The model, which is pretrained on Places365~\cite{zhou2017places}, uses a DenseNet-161~\cite{huang2017densely} architecture.
    \item The speech expert is produced using the Google Cloud API (to transcribe the speech content).
    \item The OCR expert is a word2vec encoding~\cite{mikolov2013efficient} of text detected in frames using \cite{liu2018synthetically,shi2017end}.
    
\end{itemize}

\subsection{Experts refinement -- Modifying the Kinetics action recognition model}

Apart from dropping the OCR and face experts as described in the main paper, one small modification we propose to the expert selection made by~\cite{liu2019use} is to replace the \textit{Action(IG)} from an I3D model~\cite{carreira2017quo} to an R2P1D model~\cite{tran2018closer} (matching the architecture \textit{Action(IG)}) which has also been pretrained on IG-65m~\cite{ghadiyaram2019large} and then finetuned on the Kinetics dataset~\cite{carreira2017quo}.

\section{Text embeddings description}
\label{supp:text-embed}
We use several text embeddings. In addition to the Sec. 3 from the main paper, further technical details about each of them are given below:

\begin{itemize}
    \item \textbf{mt\_grovle}~\cite{burns2019language} is a
    \say{vision-sensitive} language embedding which is adapted
    from w2v using WordNet and an original vision-language graph
    built from Visual Genome~\cite{krishna2017visual}. The size of the final pre-trained embedding is 300.
    \item \textbf{OpenAI-GPT}~\cite{radford2018improving} is a
    pre-trained text embedding which uses
    transformers~\cite{vaswani2017attention} and language modeling
    on a large corpus (the Toronto Book Corpus) (the final model has 110M params). The size of the final pre-trained embedding is 768.
    \item \textbf{RoBERTa}~\cite{liu2019roberta} is a
    BERT-based embedding~\cite{devlin2019bert}. The model is
    trained longer with bigger batch size on more data, having 125M params. The size of the final pre-trained embedding is 768.
    \item \textbf{ALBERT}~\cite{lan2019albert} is a lightweight
    modification to BERT~\cite{devlin2019bert} which overcomes
    some memory limitations, having 11M params. The size of the final pre-trained embedding is 768.
    \item \textbf{GPT2-large}~\cite{radford2019language} is a transformer-based~\cite{vaswani2017attention}
    model trained on even more data (40 Gb of text)
    without any supervision, having 774M params. The size of the final pre-trained embedding is 1280.
    \item \textbf{GPT2-xl}~\cite{radford2019language}
    is similar to GPT2-large, but has more parameters (1558M params). The size of the final pre-trained embedding is 1600.
    \item \textbf{W2V}~\cite{mikolov2013efficient} is one
    of the most popular text embeddings used in vision
    tasks. It uses a neural network model to learn
    word representations. The size of the final pre-trained embedding is 300.
\end{itemize}

\section{Further Details for Fig. 1}
\label{supp:teaser-details}
In Fig.1 from the main paper we highlight that the gain for a model that uses multiple text embeddings (last bar) is comparable with the gain of a model that uses multiple video modalities (middle bar), having as comparison a model that uses only one video modality (first bar). The first bar represents the CE~\cite{liu2019use} model trained with one video embedding, namely \textbf{Obj(IG)} (the performance of the model is 19.8±0.1 in geometric mean of R1-R5-R10). The second bar represents a CE model using 7 video modalities both for inference and training (the performance of the model is 24.4±0.1 in geometric mean of R1-R5-R10).  In the third and final bar of the chart we present the performance of using three different text embeddings with \methodName{} at training, while using only one text embedding at inference time (the performance of the model is 30.4±0.0 in geometric mean of R1-R5-R10). All the numbers are presented after the modification of the pre-processing pipeline (please see Sec.~\ref{supp:correction} for further details). All the experts used by CE~\cite{liu2019use} are described in Sec.\ref{supp:experts}.

\section{Optimization setup}
\label{supp:opt-details}
CE+ models are trained in Pytorch~\cite{paszke2019pytorch} using the Adam optimizer~\cite{kingma2014adam}. We use a learning rate of 0.001 and weight decay of 1E-5.
When using a base architecture different to the proposed CE+, we use the same hyper-parameters as in the public codebase for the underlying method (CE\footnote{\url{https://shorturl.at/ksxIS}} and MMT\footnote{\url{https://github.com/gabeur/mmt}}). For MoEE, we use the re-implementation provided by the authors of the CE method~\cite{liu2019use}.

\section{Modification to pre-processing pipeline}
\label{supp:correction}
During our preliminary analysis, we found out that some pretrained expert models produce embeddings that are fairly sensitive to jpeg compression artifacts. To address this, we re-extracted features from video frames densely extracted with minimal jpeg compression (corresponding to the use of ffmpeg~\cite{ffmpeg} and the \texttt{-qscale:v 2} flag). In order to be fair in our comparisons, we apply this corrections everywhere. Due to this factor, we re-train MoEE~\cite{miech2018learning} and CE~\cite{liu2019use} and report higher numbers. 

\section{Dataset details}
\label{supp:dataset}
To provide an extensive comparison we test our approach on seven video datasets that have been explored in recent works as benchmarks for the task of text-video retrieval. Next, we  give details about all the datasets used.\\
\indent \textbf{MSRVTT}~\cite{xu2016msr} contains 10k videos, each having 20 captions. In order to test the retrieval performance, we report results on the official split which contains 2990 videos for the test split and 497 for validation, following the setup used in~\cite{liu2019use}. We perform most of our ablations on this split. To enable comparison with as many other methods as possible, we also report results on the 1k-A split as used in~\cite{liu2019use,gabeur2020multi,patrick2020support}. For this split, we report the performance after training 100 epochs. The split contains 1000 video candidates for testing and 9000 for training. We use the same candidates as defined in~\cite{liu2019use} which are used by all the other works~\cite{yu2018joint,gabeur2020multi,patrick2020support}, using each of the 20 captions associated to each video independently during evaluation and averaging performance across them. \\
\indent \textbf{MSVD}~\cite{chen2011collecting} contains 80k English descriptions for a total of 1970 videos. We use the standard split of 1200 (training), 100 (validation) and 670 (testing) as used in other works~\cite{venugopalan2015sequence, xu2015jointly, liu2019use}. The videos from MSVD do not have audio streams. \\
\indent \textbf{DiDeMo}~\cite{anne2017localizing} contains 10464 videos sourced from a large-scale creative commons collection~\cite{thomee2016yfcc100m} and features moments of unedited, diverse content (concerts, sports, pets etc.).  The dataset comprises 3-5 pairs of descriptions per video.  We adopt the paragraph-video retrieval protocols used by~\cite{zhang2018cross,liu2019use} and use splits corresponding to 8392 train, 1065 validation and 1004 test videos.\\
\indent \textbf{LSMDC}~\cite{rohrbach2017movie} contains 118081 short video clips extracted from 202 movies. Each clip is described by a caption that is either extracted from the movie script or from transcribed DVS (descriptive video services) for the visually impaired. There are 7408 clips in the validation set and the testing is performed on 1000 videos from movies that are disjoint from the training and val sets as described in the Large Scale Movie Description Challenge (LSMDC)\footnote{\url{https://shorturl.at/cdrI6}}. \\
\indent \textbf{ActivityNet}~\cite{caba2015activitynet} contains 20k videos extracted from YouTube and has around 100K descriptive sentences. We follow the same paragraph-video retrieval setup as used in prior works~\cite{zhang2018cross, liu2019use} and report results on the \texttt{val1} split. So, we use 10009 videos for training and 4917 videos for testing. \\
\indent \textbf{VaTeX}~\cite{wang2019vatex} contains 34911 videos with multilingual captions (Chinese and English). There are 10 captions per video for each language. We follow the same protocol as in~\cite{chen2020fine, patrick2020support} and split the validation set equally (1500 validation and 1500 testing videos). In this work, we only use the English annotations.\\
\indent \textbf{QuerYD}~\cite{oncescu20queryd} contains 1815 videos in the training split and 388 and 390 for validation and testing. The videos are sourced from YouTube and cover a diverse range of visual content. The dataset contains 31441 descriptions, from which 13019 are precisely localized in the video content (having start time and end time annotations) and the other 18422 are coarsely localized. For this work, we do not use the localization annotations and report results for the official splits.

\begin{figure}
    \centering
    \includegraphics[width=\linewidth]{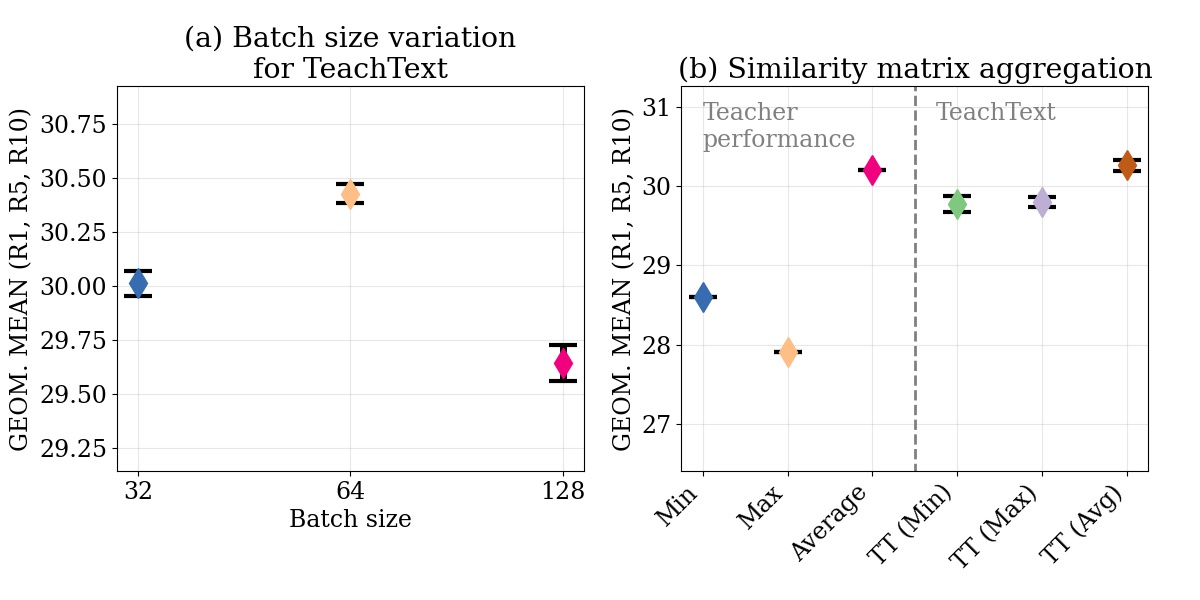}
    \caption{\textbf{(a) Batch size variation.} We vary the batch size for the MSR-VTT dataset to see how this affects the performance. We observe that batch size influences performance. The underlying architecture used for this experiment is CE+.  \textbf{(b) Similarity matrix aggregation.} We present a comparison of different similarity matrix aggregation: \textit{min}, \textit{max} and \textit{average}. As can be seen, the average aggregation has the best results (both when evaluating the teacher standalone or in conjunction with our \methodName{} algorithm).}
    \mbox{}\vspace{-0.6cm} \\
    \label{fig:supp_batch+aggregation}
\end{figure}

\section{Ablations}
\label{supp:ablations}
In this section, we present additional ablations.

\subsection{Batch size variation}
In Fig.~\ref{fig:supp_batch+aggregation}a we vary the batch size for the MSR-VTT dataset in order to see how the performance is affected. As can be seen, we obtain the best value using the same batch size as for the method without applying \methodName{} algorithm (64 in this case). 

\subsection{Similarity matrix aggregation study}
In Fig.~\ref{fig:supp_batch+aggregation}b we present several similarity matrix aggregation possibilities: \textit{min}, \textit{max} and \textit{average}. We observe that using the mean of the similarity matrices is more effective. Because of this, we use the mean as the final aggregation technique in our \methodName{} algorithm.

\subsection{Denoising}

\begin{figure}
    \centering
    \includegraphics[width=\linewidth]{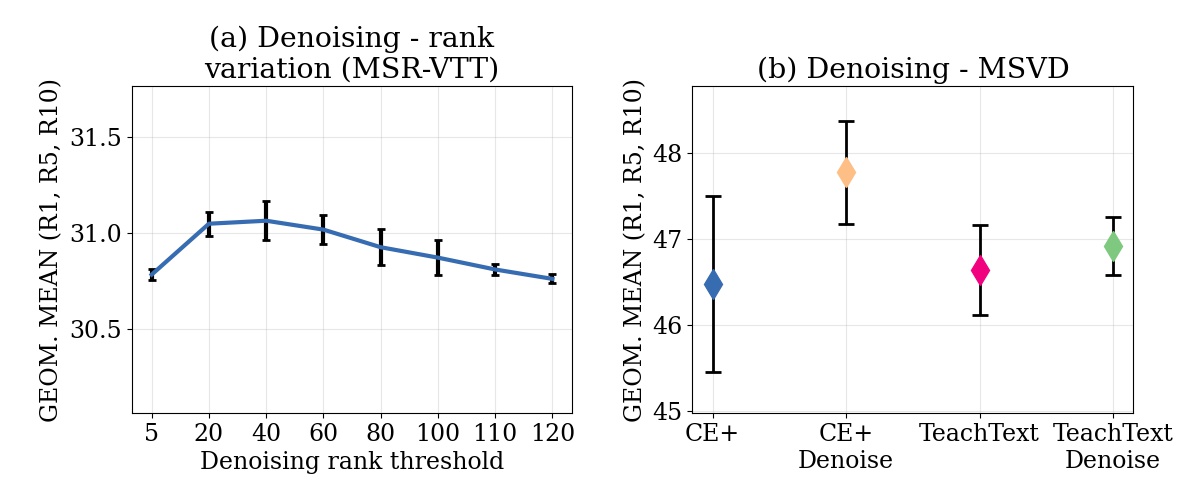}
    \caption{\textbf{(a) Rank variation for denoising.} The denoising involves dropping captions that are assigned a low ranking by the teacher for the training set. In this experiment, we vary the rank below which we drop sentences. Please note that for a rank of 5 (on the training set) the amount of dropped sentences is approximately 46\%. Note that MSR-VTT has 20 captions per video, so after applying this filter we keep on average 10 captions per video. \textbf{(b) Denoising.} We present the effect of denoising on retrieval performance on MSVD. Some of the captions available in datasets with multiple captions per video may be noisy and actively harm the training process. We estimate the degree of noise present in a caption by looking at the teacher rank and drop the caption if necessary. We observe the effectiveness of denoising when applied in isolation (\textit{CE+} vs \textit{CE+ Denoise}) and in conjunction with the full \methodName{} method. The experiment is presented for dropping sentences that rank higher than rank 100.}
    \mbox{}\vspace{-0.6cm} \\
    \label{fig:supp-denoising}
\end{figure}

In Fig.~\ref{fig:supp-denoising}a we vary the threshold used to filter out sentences from the training set. As can be seen, this denoising method is effective and it can provide a significant gain in performance. In this experiment we have found out that the best threshold for MSRVTT is rank 40. Additionally, we present denoising results in Fig.~\ref{fig:supp-denoising}b for the MSVD dataset using the 100 threshold. This method turns out to be effective in reducing noise for retrieval datasets. Denoising is not use in any other ablation studies. The final results when comparing with other state of the art methods are presented using denoising on MSRVTT and MSVD datasets.

\subsection{Distillation setup}
As stated in the main paper, the distillation setup admits a number of variants. In addition to the methods presented in Fig. 6b from the main paper, in Fig.~\ref{fig:supp-distil-loss}a we present several additional comparisons. More exactly, we test our approach against a more classical distillation setup where we directly regress the embeddings given by the teacher (\textit{Embd regress}). This setup does not follow the idea of relational distillation. Additionally, we also apply the angle distillation as introduced by~\cite{park2019relational} (\textit{Relational angle}) where we use exactly the loss as in the public code\footnote{\url{https://github.com/lenscloth/RKD}}. Please note that the drop in performance as opposed to the student without distillation can be explained by some technical challenges that we encountered in order to make the angle loss compatible with the ranking loss used for this task. Last but not least, we also show that a small improvement can be obtained by using a self learning technique, where the teacher has the exact same architecture and inputs as the student.

\subsection{Loss study}
\begin{figure}
    \centering
    \includegraphics[width=\linewidth]{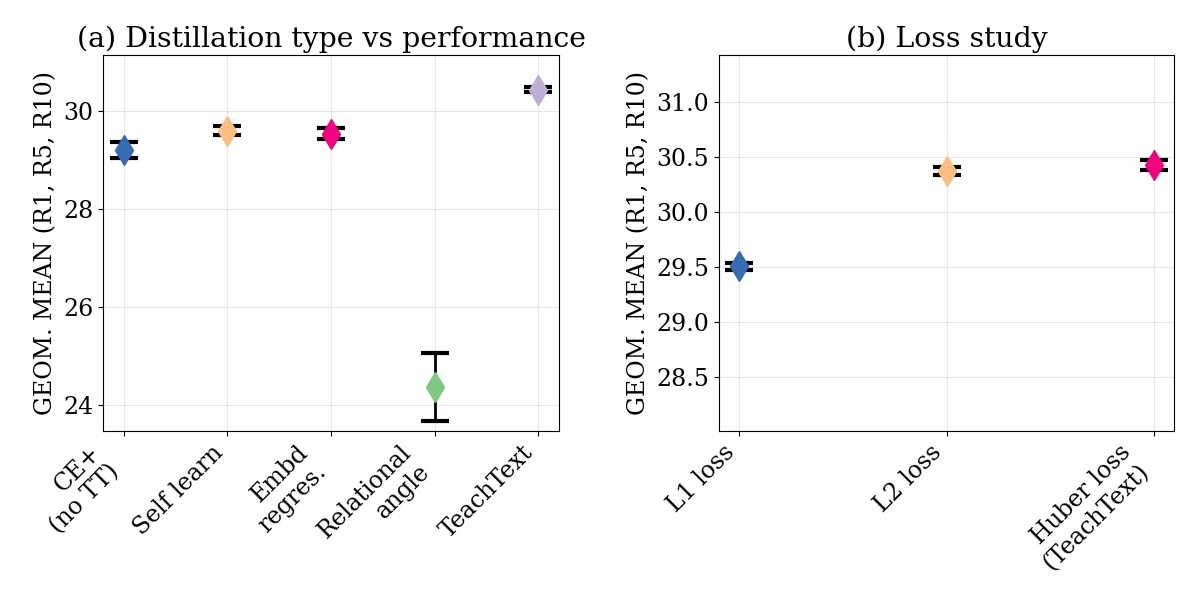}
    \caption{\textbf{(a) Distillation type.} The first bar represents the student performance without distillation (\textit{CE+}). In addition to the methods presented in the main paper, here we test other distillation approaches: \textit{Embd regress} which is a classical approach where the query and video joint embeddings are directly regressed based on the embeddings given by the teacher, \textit{Relational angle} where we apply the angle relationships as introduced by~\cite{park2019relational}. In addition, we present results of our method in a self-learning setup where the teacher is the student from a previous run (\textit{Self learn}). The last bar represent the performance of the \methodName{} approach. \textbf{(b) Loss study.} In this picture, we show how various distillation losses (L1, L2, Huber) affect the performance.}
    \mbox{}\vspace{-0.7cm} \\
    \label{fig:supp-distil-loss}
\end{figure}

In the main paper, we follow recent literature~\cite{park2019relational} and use the Huber loss for distillation. However, we wanted to see how various losses affect the performance. We test with L1 and L2 losses. As can be seen in Fig.~\ref{fig:supp-distil-loss}b, the Huber loss performs better than L1 loss and a bit better than L2.

\subsection{Mixture of architectures}
\begin{figure}
    \centering
    \includegraphics[width=\linewidth]{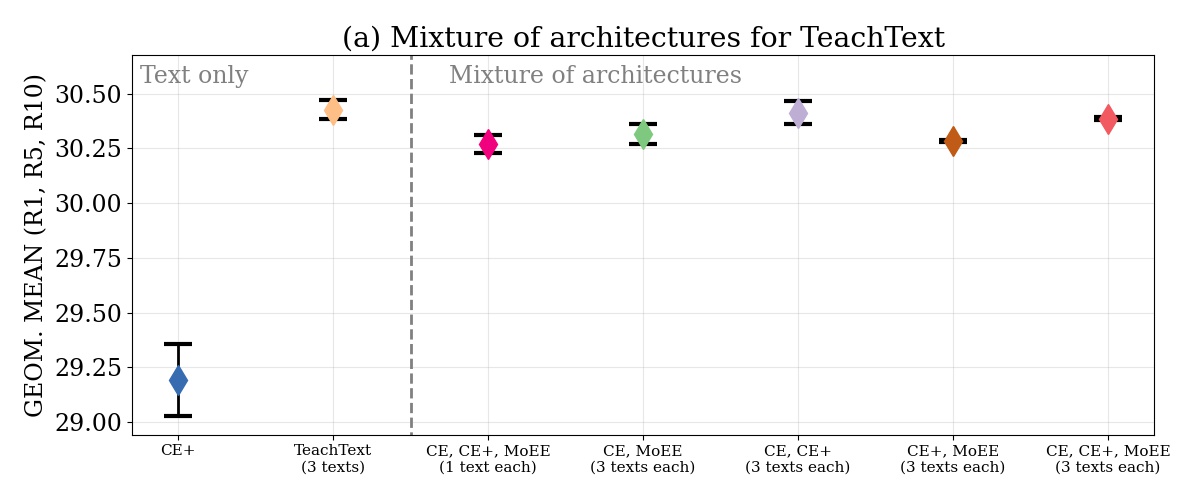}
    \caption{\textbf{Mixture of architectures.} We perform some preliminary experiments to see if the method may benefit from learning from teachers that do not share the same architecture. The x axis corresponds to the models which are used as teachers. In cases labeled with \textit{3 text each}, we used three different variations of each architecture as teachers, accounting for a total number of \textit{no. methods * 3} teachers. As can be seen, the results suggest that there is no clear benefit in using multiple architectures as teachers.}
    \mbox{}\vspace{-0.6cm} \\
    \label{fig:supp-mixture}
\end{figure}

Our \methodName{} assumes that the only difference between the teacher and the student is the used pre-trained text embedding fed to the model. However, our method is not limited to this constraint. In this section, we show how having multiple teachers, that now have a different underlying architecture affect the performance of our method. Please note that in all other ablations, the architecture is shared between student and teacher. This is the only exception. Our preliminary results shown in Fig.~\ref{fig:supp-mixture} suggest that there isn't much improvement that may be achieved by using a mixture of architectures as teachers. This is somehow expected, since these methods usually share the same video modalities so there isn't much additional information that may be captured by the combination of multiple architectures. However, we expect to get a further boost if we diversify the set of used modalities.

\subsection{Architecture extension}
\begin{table*}
\begin{center}
\resizebox{\linewidth}{!}{
\begin{tabular}{c|cc|cc|cc|cc|cc|cc}%
\hline%
\hline%
\multirow{2}{*}{Model}&\multicolumn{2}{c|}{MSRVTT}&\multicolumn{2}{c|}{MSRVTT 1k-A}&\multicolumn{2}{c|}{MSVD}&\multicolumn{2}{c|}{DiDeMo}&\multicolumn{2}{c|}{LSMDC}&\multicolumn{2}{c}{ActivityNet}\\%
&Base&\methodName{}&Base&\methodName{}&Base&\methodName{}&Base&\methodName{}&Base&\methodName{}&Base&\methodName{}\\%
\hline%
MoEE&$24.4_{\pm0.1}$&$\mathbf{25.8}_{\pm0.1}$&$41.6_{\pm0.4}$&$\mathbf{43.4}_{\pm0.6}$&$41.8_{\pm0.3}$&$\mathbf{43.2}_{\pm0.5}$&$33.2_{\pm1.4}$&$\mathbf{40.2}_{\pm0.7}$&$23.8_{\pm0.4}$&$\mathbf{26.0}_{\pm0.5}$&$40.1_{\pm0.3}$&$\mathbf{45.2}_{\pm0.1}$\\%
CE&$24.4_{\pm0.1}$&$\mathbf{25.9}_{\pm0.1}$&$42.0_{\pm0.8}$&$\mathbf{43.8}_{\pm0.3}$&$42.3_{\pm0.6}$&$\mathbf{42.6}_{\pm0.4}$&$34.2_{\pm0.4}$&$\mathbf{39.5}_{\pm0.5}$&$23.7_{\pm0.3}$&$\mathbf{25.5}_{\pm0.5}$&$40.4_{\pm0.3}$&$\mathbf{45.0}_{\pm0.6}$\\
MMT&-&-&$44.7_{\pm0.4}$&$\mathbf{45.6}_{\pm0.7}$&-&-&-&-&$24.6_{\pm0.7}$&$\mathbf{25.9}_{\pm0.6}$&$44.0_{\pm0.4}$&$\mathbf{47.9}_{\pm0.4}$\\
CE+&$29.2_{\pm0.2}$&$\mathbf{30.4}_{\pm0.0}$&$50.3_{\pm0.2}$&$\mathbf{50.9}_{\pm0.4}$&$46.5_{\pm1.0}$&$\mathbf{46.6}_{\pm0.5}$&$35.8_{\pm0.4}$&$\mathbf{40.4}_{\pm0.4}$&$28.1_{\pm0.3}$&$\mathbf{30.7}_{\pm0.3}$&$39.7_{\pm0.0}$&$\mathbf{46.3}_{\pm0.2}$\\
CE-L&$25.5_{\pm0.1}$&$\mathbf{26.9}_{\pm0.1}$&$45.7_{\pm0.2}$&$\mathbf{46.5}_{\pm0.8}$&$41.3_{\pm0.5}$&$\mathbf{42.6}_{\pm0.7}$&$36.4_{\pm0.5}$&$\mathbf{41.5}_{\pm0.4}$&$24.1_{\pm0.2}$&$\mathbf{25.9}_{\pm0.3}$&$39.6_{\pm0.5}$&$\mathbf{45.7}_{\pm0.2}$ \\
\hline%
\end{tabular}}
\end{center}
\vspace{-0.2cm}
\caption{\textbf{Method generality}. Retrieval performance on various datasets when applying \methodName{} on top of different base models. In addition to the main paper, we present results on the CE-L architecture which has a significant drop in the number of used parameters as compared to the other models. We present in bold cases where \methodName{} brings an improvement over the base architecture. As can be seen, our method is effective and brings a consistent boost independent of the base architecture.
\label{tab:supp_generality}}
\vspace{-0.1cm}
\end{table*}

\label{supp:extension}
 In addition to the main paper, we also introduce a new CE-L base architecture. This is similar to the CE~\cite{liu2019use} and CE+, but uses \texttt{w2v} as the text embedding. In this way, the number of parameters are greatly reduced, making this the most lightweight architecture in terms of number of parameters that we can create. In Tab.~\ref{tab:supp_generality}, you can see that our method \methodName{} is effective even when using this lightweight architecture. This architecture also has the lowest numbers of parameters when compared to other state of the art methods as can be seen in Tab.\ref{tab:supp-msrvtt-final-sota},\ref{tab:supp-msrvtt-jsfusion-final-sota},\ref{tab:supp-msvd-final-sota},\ref{tab:supp-didemo-final-sota},\ref{tab:supp-lsmdc-final-sota},\ref{tab:supp-activity-net-final-sota},\ref{tab:supp-vatex-final-sota},\ref{tab:supp-queryd-final-sota}.

\begin{figure}
    \includegraphics[width=\linewidth]{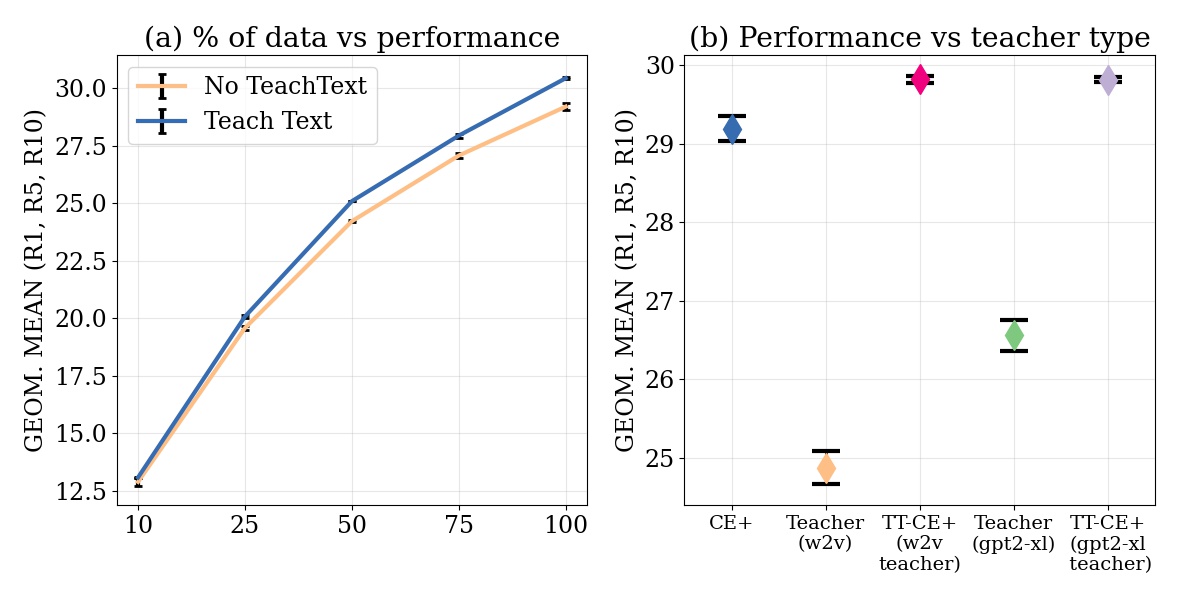}
    \caption{\textbf{(a) Amount of training data vs performance.} As it can be seen, with the increase of training data, the improvement brought by \methodName{} increases. \textbf{(b) Performance vs teacher type.} We study the influence of teachers with
    different text embeddings at input: w2v and gpt2-xl. The first point represents the performance of the student without using \methodName{}. We observe a boost in performance independent of the nature of the teacher.}
    \mbox{}\vspace{-1.5cm} \\
    \label{fig:supp_training_data+teacher_infl}
\end{figure}

\begin{figure}
    \includegraphics[width=\linewidth]{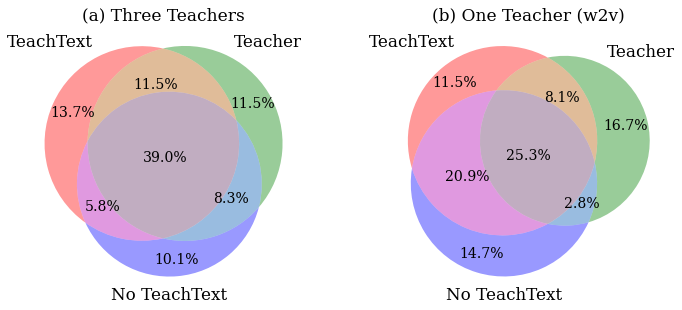}
    \caption{\textbf{Share of samples correctly retrieved samples in terms of R1 when using \methodName{} on the MSR-VTT test set.} In \textbf{sub-fig (a)} we show the case where we learn from 3 teachers, while in \textbf{sub-fig (b)} you can find the single teacher case. We can see that the model with \methodName{}, preserves most of the knowledge from the student without \methodName{}, but also acquires new information from the teacher (yellow area). Best viewed in color.
    }
    \mbox{}\vspace{-1cm} \\
    \label{fig:venn_distil}
\end{figure}

\begin{figure*}[!h]
    \centering
    \includegraphics[width=\linewidth]{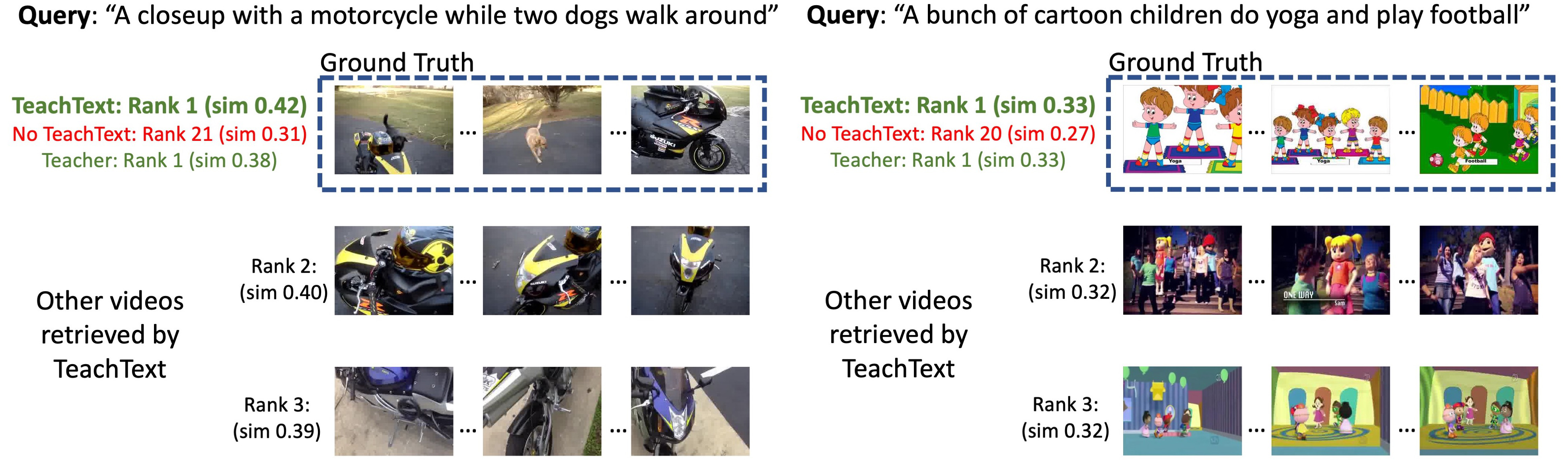}
    \caption{\textbf{Qualitative results.} We present the top 3 video retrievals for each query, given by the \methodName{} method used on top of a CE+ architecture. Moreover, we show the rank and similarity for the teacher, as well as for the student without using \methodName{} for the ground truth video. We mark in green cases where the retrieval is correct in terms of R1 and with red cases where is incorrect. For each of the cases shown, the model learns from the teacher to correct its prediction.}
    \label{fig:qualitative-example}
\end{figure*}

\subsection{Model complexity}

Changes in the pretrained text embedding strongly affect the
number of parameters. Because of this factor, using
more text embeddings at test time may strongly
affect the total number of learnable parameters available
to the model (in addition to adding the requirement to
extract additional text embeddings during inference). While the simple 'Mean' aggregation from Fig.5b
in the main paper, does not change the number of parameters,
the 'Concat' aggregation adds a significant quantity
(approx 240M learnable parameters, yielding total model
sizes of 503.98M vs 262.73M for CE+). The proposed \methodName{}
approach leaves the number of parameters untouched.

Since changing the text embedding to CE+ results in an
increase in number of learnable parameters, we also study
a CE-L architecture as a lightweight alternative in this
Suppl. Mat. (described in Sec.~\ref{supp:extension}),
which demonstrates that the gain from the proposed~\methodName{}
approach is not limited to models with many learnable
parameters. Please check Tab.\ref{tab:supp-msrvtt-final-sota},\ref{tab:supp-msrvtt-jsfusion-final-sota},\ref{tab:supp-msvd-final-sota},\ref{tab:supp-didemo-final-sota},\ref{tab:supp-lsmdc-final-sota},\ref{tab:supp-activity-net-final-sota},\ref{tab:supp-vatex-final-sota},\ref{tab:supp-queryd-final-sota} for the exact number of params for every used architecture.

\subsection{Amount of training data vs performance.}
We next study how training data quantity influences the proposed method.
In Fig.~\ref{fig:supp_training_data+teacher_infl}a we observe that by using the \methodName{} with more and more data, the performance gap increases, suggesting that its benefit may prove to be useful even in larger scale dataset scenarios.

\subsection{Teacher study}
In Fig.~\ref{fig:supp_training_data+teacher_infl}b, we study how each embedding affects the final performance. We observe that even though the model ingesting w2v embeddings has a significant lower performance than the student model without using \methodName{}, there is a significant gain when learning from the teacher which uses w2v. This again indicates that there is additional information captured by using a different text embedding which can be exploited by \methodName.

\subsection{Influence of distillation over the correctly retrieved samples}
In Fig.~\ref{fig:venn_distil} we present the shares of correctly retrieved samples in terms of R1 on the test set of the MSR-VTT dataset for the student with and without \methodName{} and for the teacher. In Fig.~\ref{fig:venn_distil}a we present results when we learn from the three teachers and in Fig.~\ref{fig:venn_distil}b we considered the case when we learn only from one teacher (namely w2v). There is a significant share of correctly retrieved sample between the student using \methodName{} and the teacher.

\section{Comparison to prior work}
\label{supp:sota}

\begin{table*}[h!]
\resizebox{\linewidth}{!}{
\begin{tabular}{c|c|c|c|c|c||c|c|c|c|c|c}%
\hline%
\hline%
Model&Task&$R@1\uparrow$&$R@5\uparrow$&$R@10\uparrow$&$MdR\downarrow$&Task&$R@1\uparrow$&$R@5\uparrow$&$R@10\uparrow$&$MdR\downarrow$&Params\\%
\hline%
Dual\cite{dong2019dual}&\texttt{t2v}&$7.7$&$22.0$&$31.8$&$32.0$&\texttt{v2t}&$13.0$&$30.8$&$43.3$&$15.0$&-\\%
HGR\cite{chen2020fine}&\texttt{t2v}&$9.2$&$26.2$&$36.5$&$24.0$&\texttt{v2t}&$15.0$&$36.7$&$48.8$&$11.0$&-\\%
MoEE\cite{miech2018learning}\footref{hq-note}&\texttt{t2v}&$11.1_{\pm0.1}$&$30.7_{\pm0.1}$&$42.9_{\pm0.1}$&$15.0_{\pm0.0}$&\texttt{v2t}&$16.5_{\pm0.1}$&$43.1_{\pm0.5}$&$57.3_{\pm0.6}$&$7.7_{\pm0.5}$&$400.41M$\\%
CE\cite{liu2019use}\footref{hq-note}&\texttt{t2v}&$11.0_{\pm0.0}$&$30.8_{\pm0.1}$&$43.3_{\pm0.3}$&$15.0_{\pm0.0}$&\texttt{v2t}&$17.0_{\pm0.5}$&$43.5_{\pm0.4}$&$57.8_{\pm0.5}$&$7.2_{\pm0.2}$&$183.45M$\\%
\hline%
TT-CE&\texttt{t2v}&$11.8_{\pm0.1}$&$32.7_{\pm0.1}$&$45.3_{\pm0.1}$&$13.0_{\pm0.0}$&\texttt{v2t}&$19.3_{\pm0.4}$&$47.0_{\pm0.7}$&$60.0_{\pm0.4}$&$6.7_{\pm0.5}$&$183.45M$\\%
TT-CE-L&\texttt{t2v}&$13.0_{\pm0.0}$&$34.6_{\pm0.1}$&$47.3_{\pm0.2}$&$12.0_{\pm0.0}$&\texttt{v2t}&$22.4_{\pm0.3}$&$50.4_{\pm0.6}$&$63.8_{\pm0.3}$&$5.3_{\pm0.5}$&$66.72M$\\%
TT-CE+&\texttt{t2v}&$\textbf{15.0}_{\pm0.1}$&$\textbf{38.5}_{\pm0.1}$&$\textbf{51.7}_{\pm0.1}$&$\textbf{10.0}_{\pm0.0}$&\texttt{v2t}&$\textbf{25.3}_{\pm0.1}$&$\textbf{55.6}_{\pm0.0}$&$\textbf{68.6}_{\pm0.4}$&$\textbf{4.0}_{\pm0.0}$&$262.73M$\\%
\hline%
\end{tabular}}
\vspace{-0.15cm}
\caption{\textbf{MSR-VTT full split: Comparison to state of the art.} \label{tab:supp-msrvtt-final-sota}}
\end{table*}

\begin{table*}[h!]
\begin{center}
\resizebox{\linewidth}{!}{
\begin{tabular}{c|c|c|c|c|c||c|c|c|c|c|c}%
\hline%
\hline%
Model&Task&$R@1\uparrow$&$R@5\uparrow$&$R@10\uparrow$&$MdR\downarrow$&Task&$R@1\uparrow$&$R@5\uparrow$&$R@10\uparrow$&$MdR\downarrow$&Params\\%
\hline%
MoEE\cite{miech2018learning}\footref{hq-note}&\texttt{t2v}&$21.6_{\pm1.0}$&$50.8_{\pm1.1}$&$65.6_{\pm0.7}$&$5.3_{\pm0.6}$&\texttt{v2t}&$22.4_{\pm0.5}$&$51.2_{\pm1.0}$&$66.1_{\pm0.4}$&$5.2_{\pm0.3}$&$400.41M$\\
CE\cite{liu2019use}\footref{hq-note}&\texttt{t2v}&$21.7_{\pm1.3}$&$51.8_{\pm0.5}$&$65.7_{\pm0.6}$&$5.0_{\pm0.0}$&\texttt{v2t}&$22.7_{\pm0.4}$&$51.8_{\pm0.4}$&$65.7_{\pm0.2}$&$5.0_{\pm0.0}$&$183.45M$\\
MMT\cite{gabeur2020multi}&\texttt{t2v}&$24.6_{\pm0.4}$&$54.0_{\pm0.2}$&$67.1_{\pm0.5}$&$4.0_{\pm0.0}$&\texttt{v2t}&$24.4_{\pm0.5}$&$56.0_{\pm0.9}$&$67.8_{\pm0.3}$&$4.0_{\pm0.0}$&$133.36M$\\%
SSB\cite{patrick2020support}&\texttt{t2v}&$27.4$&$56.3$&$67.7$&$\mathbf{3.0}$&\texttt{v2t}&$26.6$&$55.1$&$67.5$&$\mathbf{3.0}$&$-$\\%
\hline%
TT-MMT&\texttt{t2v}&$24.8_{\pm0.2}$&$55.9_{\pm0.7}$&$68.5_{\pm1.0}$&$4.3_{\pm0.5}$&\texttt{v2t}&$25.1_{\pm1.0}$&$57.1_{\pm0.8}$&$69.9_{\pm1.1}$&$4.0_{\pm0.0}$&$133.36M$\\%
TT-CE-L&\texttt{t2v}&$26.5_{\pm0.4}$&$58.0_{\pm0.8}$&$71.1_{\pm0.4}$&$4.0_{\pm0.0}$&\texttt{v2t}&$27.6_{\pm0.8}$&$58.0_{\pm0.6}$&$70.0_{\pm0.5}$&$4.0_{\pm0.0}$&$66.72M$\\%
TT-CE+&\texttt{t2v}&$\mathbf{29.6}_{\pm0.3}$&$\mathbf{61.6}_{\pm0.5}$&$\mathbf{74.2}_{\pm0.3}$&$\mathbf{3.0}_{\pm0.0}$&\texttt{v2t}&$\mathbf{32.1}_{\pm0.5}$&$\mathbf{62.7}_{\pm0.5}$&$\mathbf{75.0}_{\pm0.2}$&$\mathbf{3.0}_{\pm0.0}$&$262.73M$\\%
\hline%
\end{tabular}}
\end{center}
\vspace{-0.2cm}
\caption{\textbf{MSR-VTT 1k-A split\cite{yu2018joint}: Comparison with others.} \label{tab:supp-msrvtt-jsfusion-final-sota}}
\end{table*}

\begin{table*}[h!]
\begin{center}
\resizebox{\linewidth}{!}{
\begin{tabular}{c|c|c|c|c|c||c|c|c|c|c|c}%
\hline%
\hline%
Model&Task&$R@1\uparrow$&$R@5\uparrow$&$R@10\uparrow$&$MdR\downarrow$&Task&$R@1\uparrow$&$R@5\uparrow$&$R@10\uparrow$&$MdR\downarrow$&Params\\%
\hline%
VSE++\cite{faghri2017vse++}&\texttt{t2v}&$15.4$&$39.6$&$53.0$&$9.0$&\texttt{v2t}&$21.2$&$43.4$&$52.2$&$9.0$&-\\%
M-Cues\cite{mithun2018learning}&\texttt{t2v}&$20.3$&$47.8$&$61.1$&$6.0$&\texttt{v2t}&$\mathbf{31.5}$&$51.0$&$61.5$&$5.0$&-\\%

MoEE\cite{miech2018learning}\footref{hq-note}&\texttt{t2v}&$21.1_{\pm0.2}$&$52.0_{\pm0.7}$&$66.7_{\pm0.2}$&$5.0_{\pm0.0}$&\texttt{v2t}&$27.3_{\pm0.9}$&$55.1_{\pm1.2}$&$65.0_{\pm0.8}$&$4.3_{\pm0.5}$&$131.37M$\\%
CE\cite{liu2019use}\footref{hq-note}&\texttt{t2v}&$21.5_{\pm0.5}$&$52.3_{\pm0.8}$&$67.5_{\pm0.7}$&$5.0_{\pm0.0}$&\texttt{v2t}&$26.3_{\pm1.4}$&$53.7_{\pm0.4}$&$65.3_{\pm1.1}$&$4.8_{\pm0.2}$&$84.04M$\\%
\hline
TT-CE&\texttt{t2v}&$22.1_{\pm0.4}$&$52.2_{\pm0.5}$&$67.2_{\pm0.6}$&$5.0_{\pm0.0}$&\texttt{v2t}&$26.0_{\pm0.4}$&$53.3_{\pm0.4}$&$63.9_{\pm0.1}$&$4.9_{\pm0.1}$&$84.04M$\\%
TT-CE-L&\texttt{t2v}&$22.5_{\pm0.0}$&$53.7_{\pm0.3}$&$68.7_{\pm0.5}$&$5.0_{\pm0.0}$&\texttt{v2t}&$25.6_{\pm0.2}$&$\mathbf{55.7}_{\pm0.9}$&$65.9_{\pm0.5}$&$\mathbf{4.0}_{\pm0.0}$&$27.78M$\\%
TT-CE+&\texttt{t2v}&$\mathbf{25.4}_{\pm0.3}$&$\mathbf{56.9}_{\pm0.4}$&$\mathbf{71.3}_{\pm0.2}$&$\mathbf{4.0}_{\pm0.0}$&\texttt{v2t}&$27.1_{\pm0.4}$&$55.3_{\pm1.0}$&$\mathbf{67.1}_{\pm0.2}$&$\mathbf{4.0}_{\pm0.0}$&$87.79M$\\%
\hline%
\end{tabular}}
\end{center}
\vspace{-0.2cm}
\caption{\textbf{MSVD: Comparison to state of the art methods}. \label{tab:supp-msvd-final-sota}}
\end{table*}

\begin{table*}[h!]
\begin{center}
\resizebox{\linewidth}{!}{
\begin{tabular}{c|c|c|c|c|c|c||c|c|c|c|c|c|c}%
\hline%
\hline%
Model&Task&$R@1\uparrow$&$R@5\uparrow$&$R@10\uparrow$&$R@50\uparrow$&$MdR\downarrow$&Task&$R@1\uparrow$&$R@5\uparrow$&$R@10\uparrow$&$R@50\uparrow$&$MdR\downarrow$&Params\\%
\hline%
S2VT\cite{venugopalan2014translating}&\texttt{t2v}&$11.9$&$33.6$&-&$76.5$&$13.0$&\texttt{v2t}&$13.2$&$33.6$&-&$76.5$&$15.0$&-\\%
FSE\cite{zhang2019lookahead}&\texttt{t2v}&$13.9_{\pm0.7}$&$36.0_{\pm0.8}$&-&$78.9_{\pm1.6}$&$11.0_{\pm0.0}$&\texttt{v2t}&$13.1_{\pm0.5}$&$33.9_{\pm0.4}$&-&$78.0_{\pm0.8}$&$12.0_{\pm0.0}$&-\\%
MoEE\cite{miech2018learning}\footref{hq-note}&\texttt{t2v}&$16.1_{\pm1.0}$&$41.2_{\pm1.6}$&$55.2_{\pm1.6}$&$81.7_{\pm1.4}$&$8.3_{\pm0.5}$&\texttt{v2t}&$16.0_{\pm1.5}$&$41.7_{\pm1.9}$&$54.6_{\pm1.7}$&$81.0_{\pm1.4}$&$8.7_{\pm0.9}$&$107.26M$\\%
CE\cite{liu2019use}\footref{hq-note}&\texttt{t2v}&$17.1_{\pm0.9}$&$41.9_{\pm0.2}$&$56.0_{\pm0.5}$&$83.4_{\pm0.7}$&$8.0_{\pm0.0}$&\texttt{v2t}&$17.1_{\pm0.1}$&$41.8_{\pm0.9}$&$55.2_{\pm1.0}$&$83.0_{\pm0.8}$&$7.7_{\pm0.5}$&$79.29M$\\%
\hline
TT-CE&\texttt{t2v}&$21.0_{\pm0.6}$&$47.5_{\pm0.9}$&$61.9_{\pm0.5}$&$86.4_{\pm0.8}$&$6.0_{\pm0.0}$&\texttt{v2t}&$20.3_{\pm0.6}$&$46.6_{\pm0.6}$&$59.8_{\pm1.2}$&$85.7_{\pm0.6}$&$6.7_{\pm0.5}$&$79.29M$\\%
TT-CE-L&\texttt{t2v}&$\mathbf{22.3}_{\pm0.2}$&$\mathbf{50.1}_{\pm0.9}$&$\mathbf{64.3}_{\pm0.5}$&$\mathbf{86.9}_{\pm0.4}$&$\mathbf{5.3}_{\pm0.5}$&\texttt{v2t}&$\mathbf{21.3}_{\pm0.4}$&$\mathbf{48.3}_{\pm0.5}$&$\mathbf{62.5}_{\pm0.3}$&$86.6_{\pm0.1}$&$\mathbf{6.0}_{\pm0.0}$&$43.51M$\\%
TT-CE+&\texttt{t2v}&$21.6_{\pm0.7}$&$48.6_{\pm0.4}$&$62.9_{\pm0.6}$&$86.8_{\pm0.3}$&$6.0_{\pm0.0}$&\texttt{v2t}&$21.1_{\pm0.2}$&$47.3_{\pm0.2}$&$61.1_{\pm0.4}$&$\mathbf{86.7}_{\pm0.2}$&$6.3_{\pm0.5}$&$99.51M$\\%

\hline%
\end{tabular}}
\end{center}
\vspace{-0.2cm}
\caption{\textbf{DiDeMo: Comparison to state of the art methods}. \label{tab:supp-didemo-final-sota}}
\end{table*}

\begin{table*}[ht!]
\begin{center}
\resizebox{\linewidth}{!}{
\begin{tabular}{c|c|c|c|c|c||c|c|c|c|c|c}%
\hline%
\hline%
Model&Task&$R@1\uparrow$&$R@5\uparrow$&$R@10\uparrow$&$MdR\downarrow$&Task&$R@1\uparrow$&$R@5\uparrow$&$R@10\uparrow$&$MdR\downarrow$&Params\\%
\hline%
JSFus\cite{yu2018joint}&\texttt{t2v}&$9.1$&$21.2$&$34.1$&$36.0$&\texttt{v2t}&-&-&-&-&-\\%
MoEE\cite{miech2018learning}\footref{hq-note}&\texttt{t2v}&$12.1_{\pm0.7}$&$29.4_{\pm0.8}$&$37.7_{\pm0.2}$&$23.2_{\pm0.8}$&\texttt{v2t}&$11.9_{\pm0.5}$&$28.0_{\pm0.5}$&$37.4_{\pm0.5}$&$25.5_{\pm1.5}$&$159.78M$\\%
CE\cite{liu2019use}\footref{hq-note}&\texttt{t2v}&$12.4_{\pm0.7}$&$28.5_{\pm0.8}$&$37.9_{\pm0.6}$&$21.7_{\pm0.6}$&\texttt{v2t}&$11.4_{\pm0.4}$&$28.4_{\pm0.7}$&$36.5_{\pm0.5}$&$25.0_{\pm0.8}$&$116.86M$\\%
MMT\cite{gabeur2020multi}&\texttt{t2v}&$13.2_{\pm0.4}$&$29.2_{\pm0.8}$&$38.8_{\pm0.9}$&$21.0_{\pm1.4}$&\texttt{v2t}&$12.1_{\pm0.1}$&$29.3_{\pm1.1}$&$37.9_{\pm1.1}$&$22.5_{\pm0.4}$&$133.16M$\\%
\hline%
TT-MMT&\texttt{t2v}&$13.6_{\pm0.5}$&$31.2_{\pm0.4}$&$40.8_{\pm0.5}$&$17.7_{\pm0.5}$&\texttt{v2t}&$12.5_{\pm0.3}$&$31.3_{\pm0.6}$&$41.0_{\pm1.1}$&$18.7_{\pm0.5}$&$133.16M$\\%
TT-CE-L&\texttt{t2v}&$14.2_{\pm0.2}$&$30.6_{\pm0.3}$&$40.0_{\pm0.5}$&$20.3_{\pm0.5}$&\texttt{v2t}&$13.6_{\pm0.3}$&$30.8_{\pm0.9}$&$38.9_{\pm0.8}$&$21.5_{\pm0.4}$&$87.22M$\\%
TT-CE+&\texttt{t2v}&$\mathbf{17.2}_{\pm0.4}$&$\mathbf{36.5}_{\pm0.6}$&$\mathbf{46.3}_{\pm0.3}$&$\mathbf{13.7}_{\pm0.5}$&\texttt{v2t}&$\mathbf{17.5}_{\pm0.6}$&$\mathbf{36.0}_{\pm1.2}$&$\mathbf{45.0}_{\pm0.5}$&$\mathbf{14.3}_{\pm0.9}$&$388.24M$\\%
\hline%
\end{tabular}}
\end{center}
\vspace{-0.2cm}
\caption{\textbf{LSMDC: Comparison to state of the art methods}. \label{tab:supp-lsmdc-final-sota}}
\end{table*}

\begin{table*}[t]
\begin{center}
\resizebox{\linewidth}{!}{
\begin{tabular}{c|c|c|c|c|c||c|c|c|c|c|c}%
\hline%
\hline%
Model&Task&$R@1\uparrow$&$R@5\uparrow$&$R@50\uparrow$&$MdR\downarrow$&Task&$R@1\uparrow$&$R@5\uparrow$&$R@50\uparrow$&$MdR\downarrow$&Params\\%
\hline%
MoEE\cite{miech2018learning}\footref{hq-note}&\texttt{t2v}&$19.7_{\pm0.3}$&$50.0_{\pm0.5}$&$92.0_{\pm0.2}$&$5.3_{\pm0.5}$&\texttt{v2t}&$18.3_{\pm0.5}$&$48.3_{\pm0.8}$&$92.0_{\pm0.2}$&$6.0_{\pm0.0}$&$330.42M$\\%
CE\cite{liu2019use}\footref{hq-note}&\texttt{t2v}&$19.9_{\pm0.3}$&$50.1_{\pm0.7}$&$92.2_{\pm0.6}$&$5.3_{\pm0.5}$&\texttt{v2t}&$18.6_{\pm0.3}$&$48.6_{\pm0.7}$&$92.0_{\pm0.2}$&$6.0_{\pm0.0}$&$260.68M$\\%
HSE\cite{zhang2018cross}&\texttt{t2v}&$20.5$&$49.3$&$-$&$-$&\texttt{v2t}&$18.7$&$48.1$&-&-&-\\%
MMT\cite{gabeur2020multi}&\texttt{t2v}&$22.7_{\pm0.2}$&$54.2_{\pm1.0}$&$93.2_{\pm0.4}$&$5.0_{\pm0.0}$&\texttt{v2t}&$22.9_{\pm0.8}$&$54.8_{\pm0.4}$&$93.1_{\pm0.2}$&$4.3_{\pm0.5}$&$127.35M$\\%
SSB\cite{patrick2020support}&\texttt{t2v}&$\mathbf{26.8}$&$58.1$&$93.5$&$\mathbf{3.0}$&\texttt{v2t}&$\mathbf{25.5}$&$57.3$&$93.5$&$\mathbf{3.0}$&-\\%
\hline%
TT-MMT&\texttt{t2v}&$25.0_{\pm0.3}$&$\mathbf{58.7}_{\pm0.4}$&$95.6_{\pm0.2}$&$4.0_{\pm0.0}$&\texttt{v2t}&$24.4_{\pm0.1}$&$\mathbf{58.2}_{\pm0.3}$&$95.7_{\pm0.1}$&$4.0_{\pm0.0}$&$127.35M$\\%
TT-CE-L&\texttt{t2v}&$23.3_{\pm0.1}$&$56.3_{\pm0.1}$&$95.5_{\pm0.1}$&$4.0_{\pm0.0}$&\texttt{v2t}&$20.7_{\pm0.2}$&$52.8_{\pm0.2}$&$94.4_{\pm0.0}$&$5.0_{\pm0.0}$&$103M$\\%
TT-CE+&\texttt{t2v}&$23.5_{\pm0.2}$&$57.2_{\pm0.5}$&$\mathbf{96.1}_{\pm0.1}$&$4.0_{\pm0.0}$&\texttt{v2t}&$23.0_{\pm0.3}$&$56.1_{\pm0.2}$&$\mathbf{95.8}_{\pm0.0}$&$4.0_{\pm0.0}$&$376.02M$\\%
\hline%
\end{tabular}}
\end{center}
\vspace{-0.2cm}
\caption{\textbf{ActivityNet: Comparison to state of the art methods}. \label{tab:supp-activity-net-final-sota}}
\end{table*}

\begin{table*}[t]
\begin{center}
\resizebox{\linewidth}{!}{
\begin{tabular}{c|c|c|c|c|c||c|c|c|c|c|c}%
\hline%
\hline%
Model&Task&$R@1\uparrow$&$R@5\uparrow$&$R@10\uparrow$&$MdR\downarrow$&Task&$R@1\uparrow$&$R@5\uparrow$&$R@10\uparrow$&$MdR\downarrow$&Params\\%
\hline%
VSE\cite{kiros2014unifying}&\texttt{t2v}&$28.0$&$64.3$&$76.9$&$3.0$&\texttt{v2t}&$-$&$-$&$-$&$-$&$-$\\%
Dual\cite{dong2019dual}&\texttt{t2v}&$31.1$&$67.4$&$78.9$&$3.0$&\texttt{v2t}&$-$&$-$&$-$&$-$&$-$\\%
VSE++\cite{faghri2017vse++}&\texttt{t2v}&$33.7$&$70.1$&$81.0$&$2.0$&\texttt{v2t}&$-$&$-$&$-$&$-$&$-$\\%
HGR\cite{chen2020fine}&\texttt{t2v}&$35.1$&$73.5$&$83.5$&$2.0$&\texttt{v2t}&$-$&$-$&$-$&$-$&$-$\\%
SSB\cite{patrick2020support}&\texttt{t2v}&$44.6$&$81.8$&$89.5$&$\mathbf{1.0}$&\texttt{v2t}&$58.1$&$83.8$&$90.9$&$\mathbf{1.0}$&$-$\\%
CE\cite{liu2019use}&\texttt{t2v}&$47.9_{\pm0.1}$&$84.2_{\pm0.1}$&$91.3_{\pm0.1}$&$2.0_{\pm0.0}$&\texttt{v2t}&$60.7_{\pm1.0}$&$89.0_{\pm0.4}$&$94.9_{\pm0.2}$&$\mathbf{1.0}_{\pm0.0}$&$115.56M$\\%
\hline%

TT-CE&\texttt{t2v}&$49.7_{\pm0.1}$&$85.6_{\pm0.1}$&$92.4_{\pm0.1}$&$2.0_{\pm0.0}$&\texttt{v2t}&$62.1_{\pm0.2}$&$90.0_{\pm0.1}$&$95.3_{\pm0.1}$&$\mathbf{1.0}_{\pm0.0}$&$115.56M$\\%
TT-CE-L&\texttt{t2v}&$51.5_{\pm0.1}$&$86.5_{\pm0.1}$&$92.6_{\pm0.1}$&$\mathbf{1.0}_{\pm0.0}$&\texttt{v2t}&$\mathbf{65.0}_{\pm0.5}$&$90.3_{\pm0.5}$&$95.9_{\pm0.2}$&$\mathbf{1.0}_{\pm0.0}$&$55.07M$\\%
TT-CE+&\texttt{t2v}&$\mathbf{53.2}_{\pm0.2}$&$\mathbf{87.4}_{\pm0.1}$&$\mathbf{93.3}_{\pm0.0}$&$\mathbf{1.0}_{\pm0.0}$&\texttt{v2t}&$64.7_{\pm0.3}$&$\mathbf{91.5}_{\pm0.3}$&$\mathbf{96.2}_{\pm0.1}$&$\mathbf{1.0}_{\pm0.0}$&$223.1M$\\%
\hline%
\end{tabular}}
\end{center}
\vspace{-0.2cm}
\caption{\textbf{VaTeX: Comparison to state of the art methods}. \label{tab:supp-vatex-final-sota}}
\end{table*}

\begin{table*}[t]
\begin{center}
\resizebox{\linewidth}{!}{
\begin{tabular}{c|c|c|c|c|c||c|c|c|c|c|c}%
\hline%
\hline%
Model&Task&$R@1\uparrow$&$R@5\uparrow$&$R@10\uparrow$&$MdR\downarrow$&Task&$R@1\uparrow$&$R@5\uparrow$&$R@10\uparrow$&$MdR\downarrow$&Params\\%
\hline%
MoEE\cite{miech2018learning}&\texttt{t2v}&$11.6_{\pm1.3}$&$30.2_{\pm3.0}$&$43.2_{\pm3.1}$&$14.2_{\pm1.6}$&\texttt{v2t}&$13.0_{\pm3.1}$&$30.9_{\pm2.0}$&$43.0_{\pm2.8}$&$14.5_{\pm1.8}$&$57.75M$\\%
CE\cite{liu2019use}&\texttt{t2v}&$13.9_{\pm0.8}$&$37.6_{\pm1.2}$&$48.3_{\pm1.4}$&$11.3_{\pm0.6}$&\texttt{v2t}&$13.7_{\pm0.7}$&$35.2_{\pm2.7}$&$46.9_{\pm3.2}$&$12.3_{\pm1.5}$&$30.82M$\\%
\hline%
TT-CE&\texttt{t2v}&$14.2_{\pm1.4}$&$36.6_{\pm2.0}$&$\mathbf{51.1}_{\pm2.1}$&$\mathbf{9.7}_{\pm1.2}$&\texttt{v2t}&$14.1_{\pm0.5}$&$34.8_{\pm3.0}$&$\mathbf{49.1}_{\pm0.3}$&$\mathbf{11.3}_{\pm0.5}$&$30.82M$\\%
TT-CE+&\texttt{t2v}&$\mathbf{14.4}_{\pm0.5}$&$\mathbf{37.7}_{\pm1.7}$&$50.9_{\pm1.6}$&$9.8_{\pm1.0}$&\texttt{v2t}&$\mathbf{14.3}_{\pm0.6}$&$\mathbf{36.3}_{\pm0.9}$&$48.3_{\pm1.2}$&$\mathbf{11.3}_{\pm0.5}$&$30.82M$\\%
\hline%
\end{tabular}}
\end{center}
\vspace{-0.2cm}
\caption{\textbf{QuerYD: Comparison to state of the art methods}. \label{tab:supp-queryd-final-sota}}
\vspace{7cm}
\end{table*}

In Tab.\ref{tab:supp-msrvtt-final-sota},\ref{tab:supp-msrvtt-jsfusion-final-sota},\ref{tab:supp-msvd-final-sota},\ref{tab:supp-didemo-final-sota},\ref{tab:supp-lsmdc-final-sota},\ref{tab:supp-activity-net-final-sota},\ref{tab:supp-vatex-final-sota},\ref{tab:supp-queryd-final-sota} we make an extensive comparison of our method with other methods from the literature. In addition to the numbers reported in the main paper, we also report results for the \texttt{v2t} task. Moreover, we present the number of parameters of each method where available. As can be seen, our \methodName{} algorithm brings a clear improvement and the total number of parameters remains the same as for the base architecture. In addition to the main paper, we also introduce a new CE-L base architecture. This is similar to the CE~\cite{liu2019use} and CE+, but uses \texttt{w2v} as text embedding. In this way, the number of parameters is greatly reduced. 
As can be seen from the tables, this lightweight architecture combined with our \methodName{} algorithm has very good results 
showcasing the effectiveness of \methodName{} across different parameter regimes. Moreover, some qualitative results can be seen in Fig.\ref{fig:qualitative-example}.

\end{document}